\pdfoutput=1

\documentclass[11pt]{article}

\usepackage[final]{acl}

\usepackage{times}
\usepackage{latexsym}

\usepackage[T1]{fontenc}

\usepackage[utf8]{inputenc}

\usepackage{microtype}

\usepackage{inconsolata}

\usepackage{graphicx}
\usepackage{wrapfig}

\usepackage{hyperref}
\usepackage{url}
\usepackage{nicefrac}
\usepackage{lipsum}
\usepackage{mwe}
\usepackage{subcaption}
\usepackage{caption}

\usepackage{booktabs} 
\usepackage{array}
\usepackage{tabularx}
\usepackage{multirow}

\usepackage{algorithm, algpseudocode}
\usepackage{amsmath}

\usepackage{color, colortbl}
\definecolor{Gray}{gray}{0.95}

\title{L4Q: Parameter Efficient Quantization-Aware Fine-Tuning \\ on Large Language Models}

\author{
Hyesung Jeon$^{1}$ \quad 
Yulhwa Kim$^{2}$\thanks{Corresponding author.} \quad 
Jae-Joon Kim$^{1*}$ \\
$^{1}$Seoul National University \quad 
$^{2}$Sungkyunkwan University \\
\texttt{\{hjeon2k, kimjaejoon\}@snu.ac.kr} \quad 
\texttt{yulhwakim@skku.edu}
}

\begin{document}
\maketitle
\begin{abstract}

Due to the high memory and computational costs associated with large language models (LLMs), model compression techniques such as quantization, which reduces inference costs, and parameter-efficient fine-tuning (PEFT) methods like Low-Rank Adaptation (LoRA), which reduce training costs, have gained significant popularity. This trend has spurred active research into quantization-aware PEFT techniques, aimed at maintaining model accuracy while minimizing memory overhead during both inference and training.
Previous quantization-aware PEFT methods typically apply post-training quantization (PTQ) to pre-trained LLMs, followed by PEFT to recover accuracy loss. Meanwhile, this approach has limitations in recovering the accuracy loss.
In this paper, we propose L4Q, a method that integrates Quantization-Aware Training (QAT) with LoRA. By employing a memory-optimized layer design, L4Q significantly reduces QAT’s memory overhead, making its training cost comparable to LoRA, while preserving the advantage of QAT in producing fully quantized LLMs with high accuracy.
Our experiments demonstrate that this combined approach to quantization and fine-tuning achieves superior accuracy compared to decoupled fine-tuning schemes, particularly in 4-bit and 3-bit quantization, positioning L4Q as an efficient QAT solution. Using the LLaMA and Mistral models with instructional datasets, we showcase L4Q’s capabilities in language tasks and few-shot learning.

\end{abstract}

\section{Introduction}
\label{sec:introduction}

\begin{figure}[!ht]
    \centering
    \vspace{-10pt}
    \includegraphics[width=\linewidth]{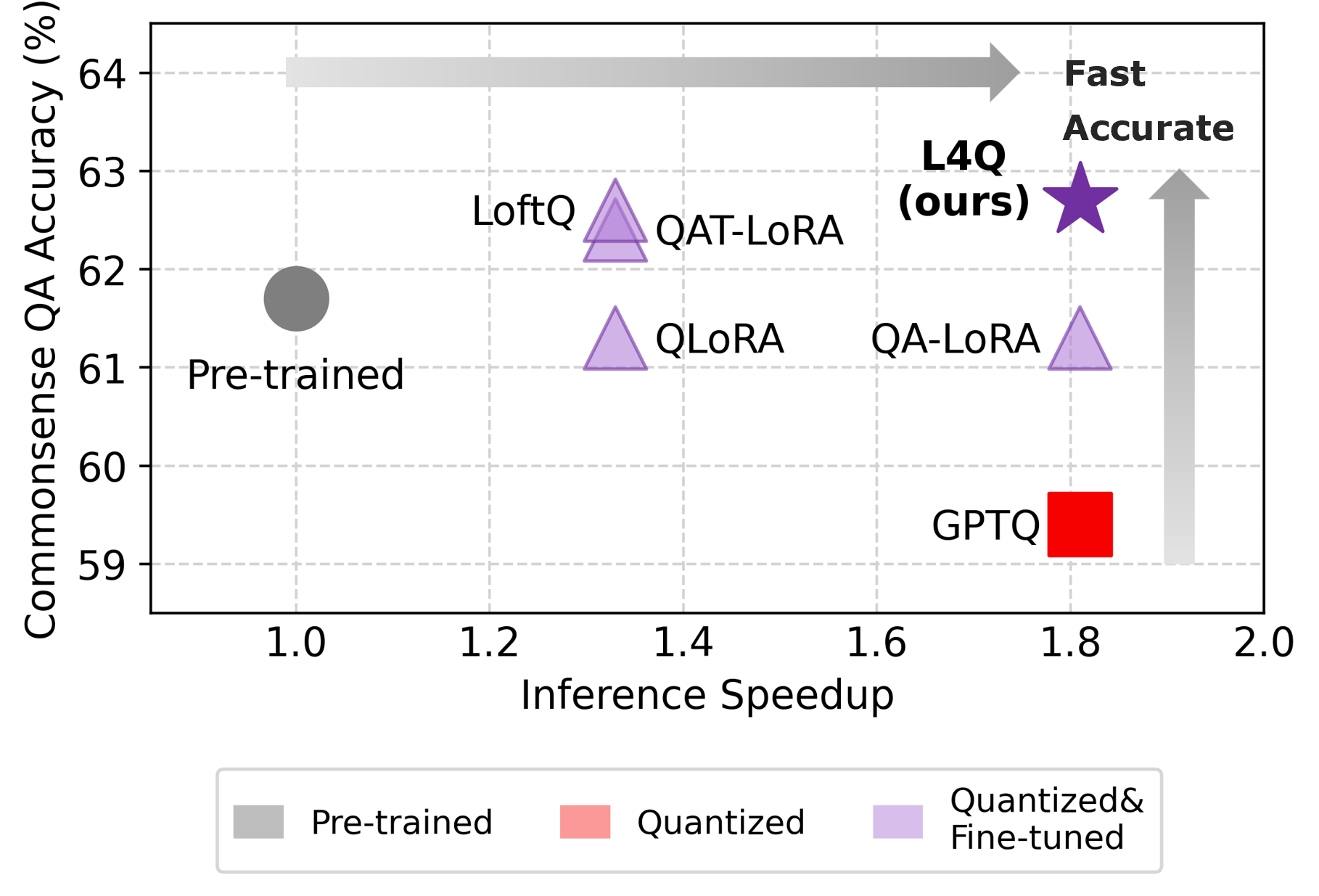}
    \vspace{-14pt}
    \caption{A diagram of accuracy and inference speedup of the quantized or fine-tuned LLaMA-1 7B models.
     L4Q produces fast and accurate quantized model.} 
    \vspace{-6pt}
    \label{fig:overview}
\end{figure}

Given their impressive scalability, Large Language Models (LLMs) such as GPT, OPT, PaLM, and LLaMA~\citep{brown2020language, zhang2022opt, chowdhery2022palm, touvron2023llama, touvron2023llama2} have become popular in natural language processing.
However, their substantial memory and computational demands pose challenges for practical deployment, making model compression~\citep{han2016deep} crucial for LLM deployment.
Quantization is a prominent method that reduces model size by lowering the bit precision of model parameters~\citep{hubara2018quantized}, so LLM quantization has been actively studied~\citep{liu2023llmqat, xiao2023smoothquant, frantar2023gptq, dettmers2023case}.
These quantization methods are generally divided into two categories: quantization-aware training (QAT) and post-training quantization (PTQ). QAT effectively reduces the quantization error by integrating quantization into the training process, where both the model weights and the quantization parameters are trained together~\citep{esser2020learned, bhalgat2020lsq}. However, applying QAT to LLM is challenging due to its significant memory overhead. As a result, PTQ, which applies quantization without retraining the entire pre-trained model weights and with minimal calibration of the quantization parameters, is widely adopted for LLM quantization~\citep{xiao2023smoothquant, lin2024awq, heo2024rethinking}.

Concurrently, to enhance the problem-solving abilities of LLMs for specific applications, fine-tuning pre-trained LLMs on downstream tasks is crucial as it improves accuracy on target tasks and related tasks~\citep{wei2022finetuned, scialom2022finetuned}.
However, fine-tuning is a resource-intensive process due to the large number of model weights involved.
Parameter-efficient fine-tuning (PEFT) addresses this issue~\citep{hu2021lora, li-liang-2021-prefix, liu2022fewshot, wang2023multitask} by training a small subset of parameters while freezing the majority of pre-trained weights.
One of the most prominent techniques within PEFT is Low-Rank Adaptation (LoRA)~\citep{hu2021lora}, which inserts trainable rank decomposition matrices into each layer to represent updates to the frozen weights.

The integration of quantization and PEFT holds significant potential for developing efficient and accurate LLMs for downstream tasks. Recent research has introduced quantization-aware PEFT approaches to achieve high-quality quantized models~\citep{dettmers2023qlora, kim2023memoryefficient, xu2023QALoRA, li2023loftq}. Previous works involve a two-stage optimization strategy: first, a PTQ technique, such as GPTQ~\citep{frantar2023gptq}, is applied to pre-trained LLMs for compression. Then, these quantized LLMs undergo PEFT, such as LoRA, where quantized weights are kept fixed and only the LoRA parameters are fine-tuned. While fine-tuning can mitigate the effects of quantization errors, separating quantization and fine-tuning into distinct stages hinders the models from achieving the best accuracy. Furthermore, as high-precision LoRA parameters are adopted alongside the quantized weight matrix, these methods eventually produce mixed-precision models, which limits the efficiency of full quantization during inference. Recently, QA-LoRA~\citep{xu2023QALoRA} addresses this issue by strictly constraining the LoRA parameter structure to integrate with quantization parameters, but this constraint can limit the fine-tuning capability.

In this paper, we propose a novel quantization-aware fine-tuning technique, named L4Q (Low-rank adaptive Learning quantization for LLMs).
L4Q addresses the limitations of PTQ-based PEFT methods by introducing QAT alongside LoRA. While QAT have advantages in reducing quantization error and LoRA enables memory-efficient training, their straightforward integration diminishes the benefits of each approach.
Therefore, L4Q carefully integrates these two approaches to properly leverage their advantages.
First, L4Q applies the quantization process after fully combining the model weights and LoRA parameters in the linear layer. This approach produces a fully-quantized model that enables memory-efficient and fast inference without limiting the training capabilities of either QAT or LoRA.
Moreover, to preserve the memory-efficient nature of LoRA during training, we design the backpropagation path of L4Q to eliminate the need to store weight gradients required for QAT.
Finally, the full integration of QAT and LoRA in the proposed L4Q allows for the joint optimization of both the quantization and LoRA parameters, thereby improving the quality of the quantized LLMs.
As a result, L4Q significantly improves the accuracy of quantized models while maintaining low memory costs during both inference and training, and achieves inference speed comparable to state-of-the-art approaches, making it a more effective solution compared to previous works, as illustrated in Figure~\ref{fig:overview}.

\section{Backgrounds}
\label{sec:backgrounds}

\subsection{PEFT with LoRA}

LoRA inserts the rank-decomposition matrices composed of a pair of parameter matrices $A \in R^{r \times i}$ and $B \in R^{o \times r}$.
Here, $i$ and $o$ represent the size of input and output dimensions of the original weight matrix, respectively. $r \ll i, o$ is the rank of the LoRA matrices, and $\alpha$ is a constant that adjusts the influence of the LoRA matrices.
During the fine-tuning process, the pre-trained weight matrix $W_{0} \in R^{o \times i}$ is frozen, preserving the pre-trained features.
For a given input activation $X \in R^{i \times s \times b}$ ($s$: sequence length, $b$: batch size), the output $Y \in R^{o \times s \times b}$ of a layer utilizing LoRA is computed as follows:
\begin{equation}
    \label{eq:lora_forward}
        Y = W_{0}X + \alpha BAX 
\end{equation}
The fine-tuning of the LoRA parameters is guided by the gradient of a loss function $L$, which is calculated with respect to each parameter matrix. The gradients are derived as follows:
\begin{equation}
    \label{eq:lora_backward}
        \frac{\partial L}{\partial A} = \alpha \frac{\partial L}{\partial \tilde{X}} X^{\top}, \quad
        \frac{\partial L}{\partial B} = \alpha \frac{\partial L}{\partial Y} \tilde{X}^{\top}
\end{equation}
Here, $\tilde{X} := AX$ represents the intermediate input activation of $B$, which is the transformation of $X$ by $A$. These gradients guide the adjustment of the LoRA parameters to minimize the loss and more accurately approximate the necessary updates to the original model weights.

\subsection{Quantization}
Uniform quantization is a widely used quantization scheme due to its simplicity and broad compatibility with various computing kernels and hardware units~\citep{liu2022uniform}.
Therefore, we refer to the term `quantization' specifically as uniform quantization throughout this paper.
A common practice is to organize a quantization group consisting of a certain number of consecutive weight elements that share the same quantization scale $s$ and zero-point (bias) $b$.

The weights $W$ within the quantization group are quantized according to the following equation:
\begin{equation}
    \label{eq:quant}
        \tilde{w} = \text{round}(\text{clamp}(\frac{W-b}{s},Q_{N},Q_{P}))
\end{equation}
Here, $\tilde{w}$ denotes the quantized integer value. Clamping is applied within the range $Q_{N} = -2^{n-1}$ to $Q_{P} = 2^{n-1} - 1$, where $n$ is the bit-width, followed by the rounding function. We also note that $W_q := \tilde{w} \times s + b$ represents the dequantized version of the quantized weight, which is adjusted using $s$ and $b$ from $\tilde{w}$ to approximate the original weight.

During QAT, the straight-through estimator (STE) approximates the derivative of the rounding function with an identity function~\citep{bengio2013estimating, choi2018pact, esser2020learned}, enabling gradients to propagate through non-differentiable rounding operations and allowing effective weight parameter training.
LSQ~\citep{esser2020learned} and LSQ+~\citep{bhalgat2020lsq} extend this process by training quantization parameters $s$ and $b$, alongside the model weights. This tuning scheme provides finer control over quantization, improving model accuracy.
The quantization parameters tuning during backpropagation proceeds by using the chain rule via $W_q$. This is denoted as follows:
\begin{equation}
        \frac{\partial L}{\partial s} =  \frac{\partial L}{\partial W_{q}}
        \frac{\partial W_{q}}{\partial s}, \quad 
        \frac{\partial L}{\partial b} =  \frac{\partial L}{\partial W_{q}}
        \frac{\partial W_{q}}{\partial b}
\end{equation}
As a consequence, the backpropagation process requires the weight gradient $\frac{\partial L}{\partial W_{q}}$ and the computation of the terms $\frac{\partial W_{q}}{\partial s}, \frac{\partial W_{q}}{\partial b}$ regarding the non-linear STE function.
\begin{equation}
    \label{eq:quant_lsq_s}
        \frac{\partial W_{q}}{\partial s} = -w + \tilde{w},
        \quad
        \mathrm{if} \enspace Q_{N}\leq w\leq Q_{P}
\end{equation}
\begin{equation}
    \label{eq:quant_lsq_b}
        \frac{\partial W_{q}}{\partial b} =  1,
        \quad
        \mathrm{if} \enspace w<Q_{N} \enspace or \enspace w>Q_{P}
\end{equation}
More details on QAT with LSQ and LSQ+ are provided in Appendix~\ref{appen:qat}.

\begin{figure*}[t]
    \centering
    \includegraphics[width=0.95\textwidth]{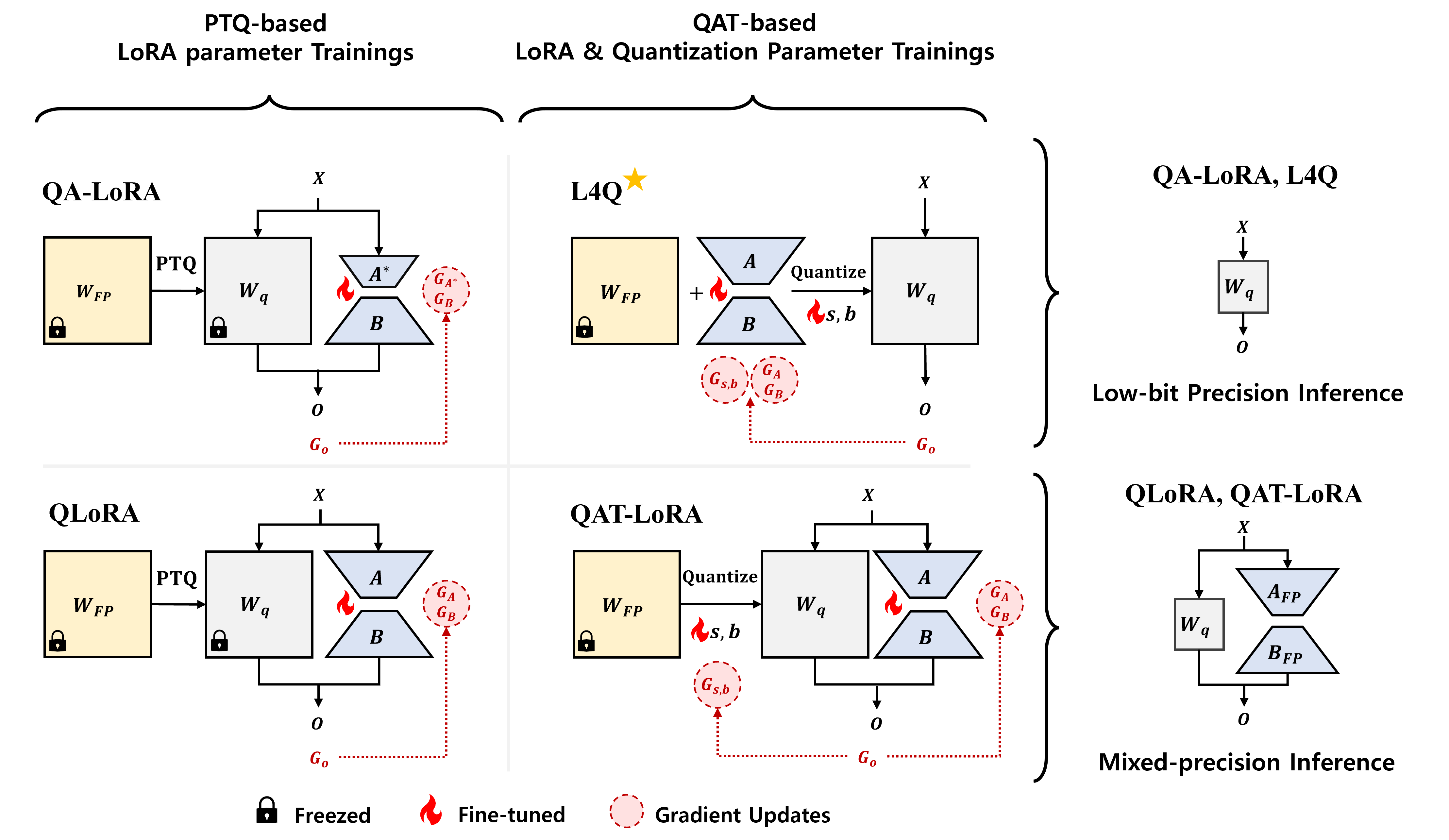}
        \caption{A categorization of training scheme and inference strategy of QA-LoRA, QLoRA, QAT-LoRA, and L4Q.
    Compared to QA-LoRA, L4Q utilizes higher optimization ability with non-constrained LoRA parameters and quantization parameters.
    Additionally, compared to QLoRA and QAT-LoRA, L4Q exploit fully-quantized weights rather than the mixed-precision weights during inference and perform a solid co-optimization of parameters.
    }
    \vspace{-5pt}
\label{fig:diagram}
\end{figure*}

\subsection{LLM Quantization}
Quantization compress LLMs by lowering the bit precision of model parameters~\citep{hubara2018quantized}.
A key challenge is the introduction of quantization errors that reduce model accuracy, leading to extensive research aimed at mitigating these losses through calibration or training.
A notable examples of PTQ for LLM compression are GPTQ~\citep{frantar2023gptq} and OmniQuant~\citep{shao2023OmniQuant}.
In contrast, QAT integrates quantization into the training process, adaptively training model parameters to account for quantization errors during training, ensuring that the quantized model retains much of its accuracy and functionality through training.
Despite its advantages, QAT faces challenges, primarily due to its high training overhead, which limits its use in resource-intensive models like LLMs. The memory overhead of QAT stems from storing weight gradients and their optimizer states, each requiring multiple times the memory of the weights. Hence, even applying QAT to a 7B model requires approximately 80GB of memory.

\subsection{Quantization-Aware PEFT}
\label{sec:backgrounds_qapeft}
To improve the accuracy of quantized LLMs, recent research has introduced quantization-aware PEFT approaches~\citep{dettmers2023qlora, kim2023memoryefficient, xu2023QALoRA, li2023loftq}. Among these, QLoRA~\citep{dettmers2023qlora}, QA-LoRA~\citep{xu2023QALoRA}, and LoftQ~\citep{li2023loftq} stand out as notable methods.
As illustrated in Figure~\ref{fig:diagram}, QLoRA begins by applying PTQ to a pre-trained model. After this initial quantization, LoRA fine-tuning is performed, with the quantized weight parameters kept frozen. This allows the method to correct quantization errors during the fine-tuning.
However, QLoRA introduces computational inefficiencies during inference due to the additional forward path on LoRA parameters. This inefficiency arises because the high-precision LoRA parameters and low-precision quantized weights cannot be merged into low-precision values.
Advanced methods~\citep{li2023loftq, qin2024lrqlora} that build upon QLoRA and share its layer structure also suffer from this issue.
We further examine the impact of this unmerged LoRA path on inference efficiency by comparing the speed of fully-quantized models with mixed-precision models in Section~\ref{sec:experiments}. 

QA-LoRA~\citep{xu2023QALoRA} addresses the issue of high-precision LoRA parameters by modifying the structure of the LoRA matrix, allowing these parameters to be integrated with the quantized weights after training (Figure~\ref{fig:diagram}). 
The input dimension of the LoRA matrix $A$ is set to the number of weight quantization groups. 
This adjustment ensures that each element of $BA$ corresponds directly with individual quantization groups, enabling the LoRA parameters to be seamlessly integrated into the quantization bias as $b' = b - \alpha BA$ at the end of training.
Hence, QA-LoRA shares the same objective as our work.
However, this solution presents a new challenge: the constrained LoRA structure in this setup limits the model's ability to achieve optimal accuracy during the PEFT stage.

A broader issue with existing quantization-aware PEFT methods is that fine-tuning begins from a pre-quantized model with inherent quantization errors, which is suboptimal compared to starting from a pre-trained model. LoftQ attempts to mitigate these errors by approximating them with LoRA using iterative singular-value decomposition (SVD). However, this approach still cannot achieve a single forward path due to the high-precision LoRA parameters, limiting subsequent adaptation.
These challenges underscore the need for further research to improve quantization-aware PEFT techniques, focusing on enhancing both quantization and PEFT processes for better accuracy and inference efficiency in LLMs.

\section{Methods}
\label{sec:methods}

\subsection{Straight Integration of QAT and LoRA}
\label{sec:qat_lora}

\textbf{QAT-LoRA} \enspace
One of the key principles in our proposed L4Q scheme is the integration of the QAT and LoRA to facilitate the simultaneous calibration for quantization and fine-tuning on downstream tasks. To achieve this, we begin with a straightforward integration of QAT and LoRA, referred to as QAT-LoRA, which serves as our baseline approach for combining QAT and PEFT.

In QAT-LoRA, pre-trained weights are frozen, while LoRA parameters are added to the linear layers (Figure~\ref{fig:diagram}). Additionally, quantization scales and bias parameters are introduced, similar to advanced QAT techniques like LSQ, which are crucial for calibrating the quantization function.
Freezing the weights reduces the need for optimizer states, while a small number of LoRA and quantization parameters are introduced to approximate updates to the weight matrix and to update the quantization function, respectively. This results in more efficient memory usage compared to standard QAT. 
Detailed analysis results of the memory efficiency of QAT-LoRA is further discussed in Section~\ref{sec:experiments}.

\noindent
\textbf{Limitations of QAT-LoRA} \enspace
While the previous section introduced a straightforward integration of QAT and LoRA—where quantized weights and LoRA parameters are maintained separately—this approach presents several limitations.
First, although freezing the pre-trained weights eliminates the need for optimizer states, weight gradients $\frac{\partial L}{\partial W_q}$ must still be stored to update the quantization parameters, as shown in Equation~\ref{eq:quant_lsq_s} and Equation~\ref{eq:quant_lsq_b}.
As a result, QAT-LoRA still incurs memory overhead from weight gradients, undermining the memory efficiency benefits of LoRA fine-tuning.
Secondly, QAT-LoRA produces a mixed-precision model with both quantized weights and high-precision LoRA parameters. This mixed-precision approach negates the advantages of LLM quantization, similar to previous methods such as QLoRA and LoftQ discussed in Section~\ref{sec:backgrounds_qapeft}.
Lastly, the gradient updates for quantization and LoRA parameters are decoupled, limiting the potential for comprehensive optimization across the model.
As outlined in Equations ~\ref{eq:quant_lsq_s} and ~\ref{eq:quant_lsq_b}, updates to the quantization parameters rely on the quantized weight matrix $W_q$, while updates to the LoRA parameters depend on weights $A$ and $B$.
This limits the effectiveness of model training, as it prevents holistic adjustments where changes in LoRA parameters could directly influence quantization adjustments and vice versa.

To address these challenges, we introduce L4Q, an enhanced integration of QAT and LoRA. L4Q features an advanced layer design that seamlessly integrates QAT and LoRA. By applying quantization after merging the weights and LoRA parameters, along with a custom backpropagation path that reduces the memory overhead from the complex quantization and LoRA processes, L4Q effectively overcomes the primary challenges encountered with QAT-LoRA.

\subsection{L4Q: Low-rank Adaptive Quantization-Aware Fine-tuning}
\label{sec:l4q}
\textbf{Fully-Quantized Linear Layer} \enspace
As high-precision LoRA weights introduces inference overhead, it is crucial to design a fully-quantized linear layer. In this context, L4Q first combines the original weights $W_0$ and the LoRA parameters $BA$ into a unified parameter matrix:
\begin{equation}
    \label{eq:l4q_Wcombine}
        W_{comb} = W_0 + \alpha BA
\end{equation}
Then, quantization is applied to the fully combined weight $W_{comb}$ and produces $\tilde{w}$ and $W_q$:
\begin{equation}
    \label{eq:l4q_Wquant}
        \enspace \tilde{w} =\text{round}(\text{clamp}(\frac{W_{comb} -b}{s},Q_{N},Q_{P}))
\end{equation}
\begin{equation}
\label{eq:l4q_Wdequant}
    W_q=\tilde{w}\times s + b
\end{equation}
In this way, during inference, L4Q only uses quantized weights $W_q$, simplifying the forward path of the linear layer from Equation~\ref{eq:lora_forward} to as follows:
\begin{equation}
    \label{eq:l4q_forward}
        Y = W_q X 
\end{equation}
While QA-LoRA also achieves fully-quantized linear layers by introducing constraints on the LoRA parameter structure, the proposed L4Q imposes no such restrictions. This flexibility allows L4Q to fully leverage the benefits of LoRA-based fine-tuning, all with fully-quantized linear layers.

\noindent
\textbf{Memory Efficient QAT} \enspace
As discussed in the previous section, QAT requires weight gradients to train quantization parameters $s$ and $b$. Since weight gradients are a major source of memory overhead during training, we compute these gradients locally in the backpropagation path as follows:
\begin{equation}
    \frac{\partial L}{\partial W_q} = \frac{\partial L}{\partial Y} X^{\top}
\end{equation}

We then use these weight gradients to calculate gradients of $s$ and $b$ with Equation~\ref{eq:quant_lsq_s} and~\ref{eq:quant_lsq_b}. Once the gradient computation for the linear layer is complete, the weight gradients are immediately flushed to conserve memory.

\noindent
\textbf{Efficient LoRA Training} \enspace
Unlike the conventional LoRA backward path which does not involve the weight $W_0$ as described in Equation~\ref{eq:lora_backward}, computing the gradients of the LoRA parameters in the L4Q linear layer follows a more complicated backpropagation path, tracing back from Equation~\ref{eq:l4q_forward} to Equation~\ref{eq:l4q_Wcombine}. 
Because a non-linear quantization function is applied after LoRA during quantization in L4Q, the gradients of the LoRA parameters depend on both the weights (in their quantized form $w$) and the weight gradients. This process can be described as follows:
\begin{equation}
    \label{eq:l4q_backward}
        \frac{\partial L}{\partial A} = \frac{\partial L}{\partial W_{q}}\frac{\partial W_{q}}{\partial A}, 
    \quad
        \frac{\partial L}{\partial B} = \frac{\partial L}{\partial W_{q}}\frac{\partial W_{q}}{\partial B}
\end{equation}
We reuse the weight gradient $\frac{\partial L}{\partial W_{q}}$ that have been computed previously for quantization parameter update. Therefore, we only compute $\frac{\partial W_{q}}{\partial A}$ and $\frac{\partial W_{q}}{\partial B}$ to obtain the gradients of LoRA parameters. 
Both terms are derived by applying the chain rule from Equation~\ref{eq:l4q_Wquant} to Equation~\ref{eq:l4q_Wcombine}.
Since Equation~\ref{eq:l4q_Wquant} contains a rounding function, we apply STE and clamping with conditional gradient propagation to $\frac{\partial W_{q}}{\partial A}$ and $\frac{\partial W_{q}}{\partial B}$.
This leads to the following expressions for $\frac{\partial W_{q}}{\partial A}$ and $\frac{\partial W_{q}}{\partial B}$:

\begin{equation}
    \label{eq:l4q_backward_wa}
        \frac{\partial W_{q}}{\partial A} = \begin{cases}
        \alpha B^{\top}, & \quad \mathrm{if} \enspace Q_{N} \leq w \leq Q_{P} \\
        0, & \quad otherwise
        \end{cases}
\end{equation}
\begin{equation}
    \label{eq:l4q_backward_wb}
        \frac{\partial W_{q}}{\partial B} = \begin{cases}
        \alpha A^{\top}, & \quad \mathrm{if} \enspace Q_{N} \leq w \leq Q_{P} \\
        0,  & \quad otherwise
        \end{cases}
\end{equation}
Therefore, the proposed L4Q efficiently processes LoRA training by simply reusing the weight gradients computed for QAT parameter training.
For more detailed explanations of the gradient calculation in L4Q, please refer to Appendix~\ref{appen:l4q}, and the memory efficiency of L4Q will be further examined in Section~\ref{sec:experiments}.

\noindent
\textbf{Joint Quantization and Low-rank Adaptation} \enspace
Since $\frac{\partial L}{\partial W_q}$ is involved in the gradient calculation for the LoRA parameters (Equation~\ref{eq:l4q_backward}), the proposed L4Q ensures that the impact of quantization is directly reflected in the updates to the LoRA parameters. This enables the joint optimization of LoRA parameters and the quantization process, enhancing the accuracy of the fully-quantized LLMs. 

In summary, the proposed L4Q produces a fully-quantized model for memory-efficient and fast LLM inference by fully integrating the model weights and LoRA parameters prior to the quantization process. Additionally, the training process of L4Q is memory-efficient due to careful handling of gradient computation for quantization. Finally, L4Q can improve the accuracy of quantized LLMs through the joint optimization of the quantization and LoRA parameters.

\subsection{Quantization Parameter Initialization}
\label{sec:quantinit}
LSQ+, a previous QAT approach, sets the quantization range based on the weight standard deviation. This is effective for CNNs, but it does not work well on LLMs because activation outliers and their corresponding salient weights are crucial for model performance~\citep{xiao2023smoothquant, dettmers2022llmint8, lin2024awq}. Hence, when initializing quantization parameters, it is important to address the outlier sensitivity of LLMs.
To address this, we propose L4Q\textsubscript{init}, a symmetric quantization scheme that minimizes clipping errors by using a conservative scale to capture both minimum and maximum outliers. The quantization scale is defined by the following equation:
\begin{equation}
    \label{eq:l4q_quant_init}
    s = Max(|\frac{Min(W)}{Qn}|, |\frac{Max(W)}{Qp}|)
\end{equation}

\begin{figure}[t]
    \centering
    \includegraphics[width=\linewidth]{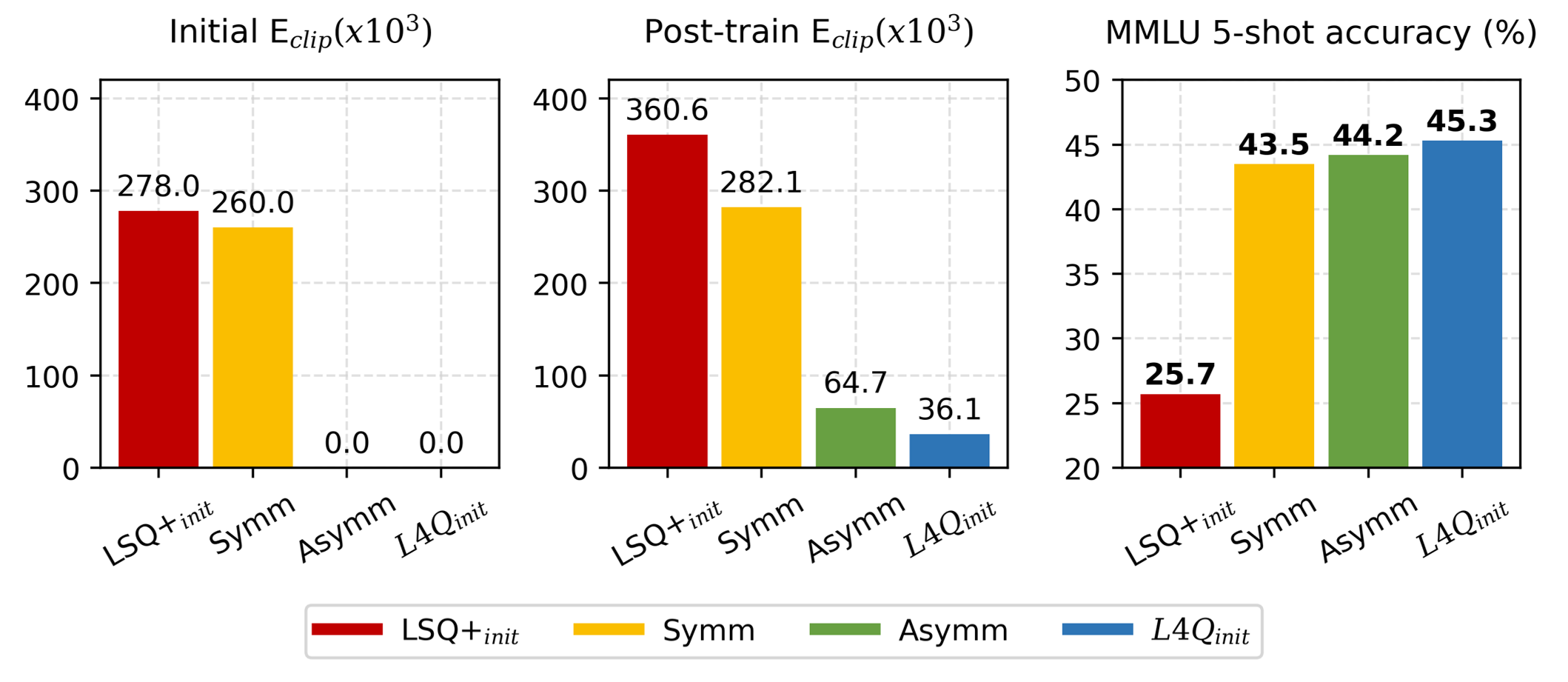}
    \caption{MMLU 5-shot results and clipping errors for LLaMA-2 7B models after 100 training steps. The results include clipping errors at both initialization and post-training for the LSQ+, symmetric, asymmetric, and L4Q initialization methods.}
    \vspace{-5mm}
\label{fig:init}
\end{figure}

We evaluate models trained with L4Q using various initialization methods, including conventional min/max-based symmetric and asymmetric schemes, and LSQ+\textsubscript{init} schemes. 
We compare the accuracy and the sum of clipping errors caused by overflowed outliers at both initialization and after training.
As shown in Figure~\ref{fig:init}, LSQ+\textsubscript{init} and symmetric initialization result in higher clipping errors. While asymmetric initialization avoids clipping initially, its tight range, defined by minimum-to-maximum values, becomes insufficient as weights evolve, leading to increased clipping errors during fine-tuning.
In contrast, L4Q\textsubscript{init} accounts for the broader weight distribution in LLMs, effectively reducing clipping errors. As a result, L4Q\textsubscript{init} achieves the highest accuracy, whereas LSQ+\textsubscript{init} struggles to recover from quantization errors.
A detailed explanation of LSQ+\textsubscript{init}, symmetric, asymmetric, and L4Q initialization methods is provided in Appendix~\ref{appen:quant_init}.

\section{Experiments}
\label{sec:experiments}
\subsection{Experimental Settings}
\textbf{Target Foundation Models} \enspace
OpenLLaMA\footnote{https://github.com/openlm-research/open\_llama} 3B model, LLaMA family models~\citep{touvron2023llama, touvron2023llama2} 7B to 33B, and Mistral-v0.1 7B~\citep{jiang2023mistral7b} model are used for the evaluation.

\noindent
\textbf{Baselines} \enspace
We compare the proposed L4Q with previous quantization methods and quantization-aware PEFT methods. The baseline quantization methods considered are LSQ for QAT, and GPTQ and OmniQuant for PTQ. For quantization-aware PEFT baselines, we include QLoRA, QA-LoRA, and LoftQ.
We apply uniform quantization in our experiments to ensure consistency across methods. The methods that were originally deployed with non-uniform quantization are denoted with an asterisk (*).
For methods with LoRA adapters, including L4Q, the adapters are initialized following the original scheme: $A$ is initialized using Kaiming-uniform, while $B$ is initialized to zeros.
We also note that the term \emph{fully-quantized} refers to a condition in which all linear layers are quantized, while non-linear components and the head layer remain in higher precision.

\noindent
\textbf{Evaluation Setups} \enspace
We establish the L4Q framework based on the open-source frameworks: Lit-GPT\footnote{https://github.com/Lightning-AI/lit-gpt.git} and huggingface transformers\footnote{https://github.com/huggingface/transformers.git}.
The models are symmetrically quantized with quantization group size of 128 for the LLaMA and Mistral models, and 64 for the OpenLLaMA models due to its small channel size. Also, the models are fine-tuned with a LoRA rank $r$ of 4 by default, and 8 for Mistral. We double the rank $r$ in QA-LoRA for a fair comparison in terms of the number of LoRA parameters. Further details on the effect of rank size and quantization group size can be found in Appendix~\ref{appen:l4q_hyperparam}.
We use the Stanford-Alpaca~\citep{alpaca}, a dataset that consists of 50k training samples and 2k validation samples generated from the GPT 3.5~\citep{brown2020language}.
We use bfloat16 precision for numerically stable fine-tuning.
All experiments are conducted on an NVIDIA A100 80GB GPU. Detailed hyperparameter settings for fine-tuning are provided in Appendix~\ref{appen:ex_settings}.

\noindent
\textbf{Evaluation Metrics} \enspace
We evaluate the accuracy of LLMs on Commonsense QA (CSQA)~\citep{leo_gao_lmeval} and MMLU~\citep{hendrycks2021measuring} benchmarks.
The CSQA benchmark comprises seven multiple-choice tasks designed to evaluate language models~\citep{zellers2019hellaswag, bisk2019piqa, clark2018arc, sakaguchi2019winogrande, clark2019boolq, talmor2019commonsenseqa}.
The MMLU benchmark spans four subject categories made up of 57 subcategories of language tasks.

\subsection{Evaluation Results}

\newcolumntype{L}[1]{>{\arraybackslash}m{#1}}
\newcolumntype{P}[1]{>{\centering\arraybackslash}m{#1}}
\begin{table}[!t]
    \resizebox{1.0\linewidth}{!}{
    \begin{tabular}{L{1.5cm}P{1.5cm}P{1.3cm}P{1.3cm}P{1.3cm}}
    \toprule
     \rowcolor{Gray}& \small OpenLLaMA & \multicolumn{3}{c}{\small LLaMA} \\
     \rowcolor{Gray}Methods & 3B & 7B & 13B & 33B \\
     \arrayrulecolor{black} \midrule
     \small LoRA  & 15.1 & 25.1 & 43.8 & 71.9 \\
     \arrayrulecolor{lightgray}\midrule
     \small QAT  & 44.2 & 79.5  & OOM & OOM \\
     \small QAT-LoRA  & 22.6 & 41.9 & 70.6 & OOM \\
     \small L4Q  & 15.3 & 25.4 & 44.3 & 73.2 \\
    \arrayrulecolor{black}\bottomrule
    \end{tabular}
    }
    \caption{Memory cost (GB) for fine-tuning LLMs on NVIDIA A100 GPU. (OOM: Out of Memory)}
    \label{table:memory}
\end{table}

\begin{figure}[t]
\centering
    \includegraphics[width=0.9\linewidth]{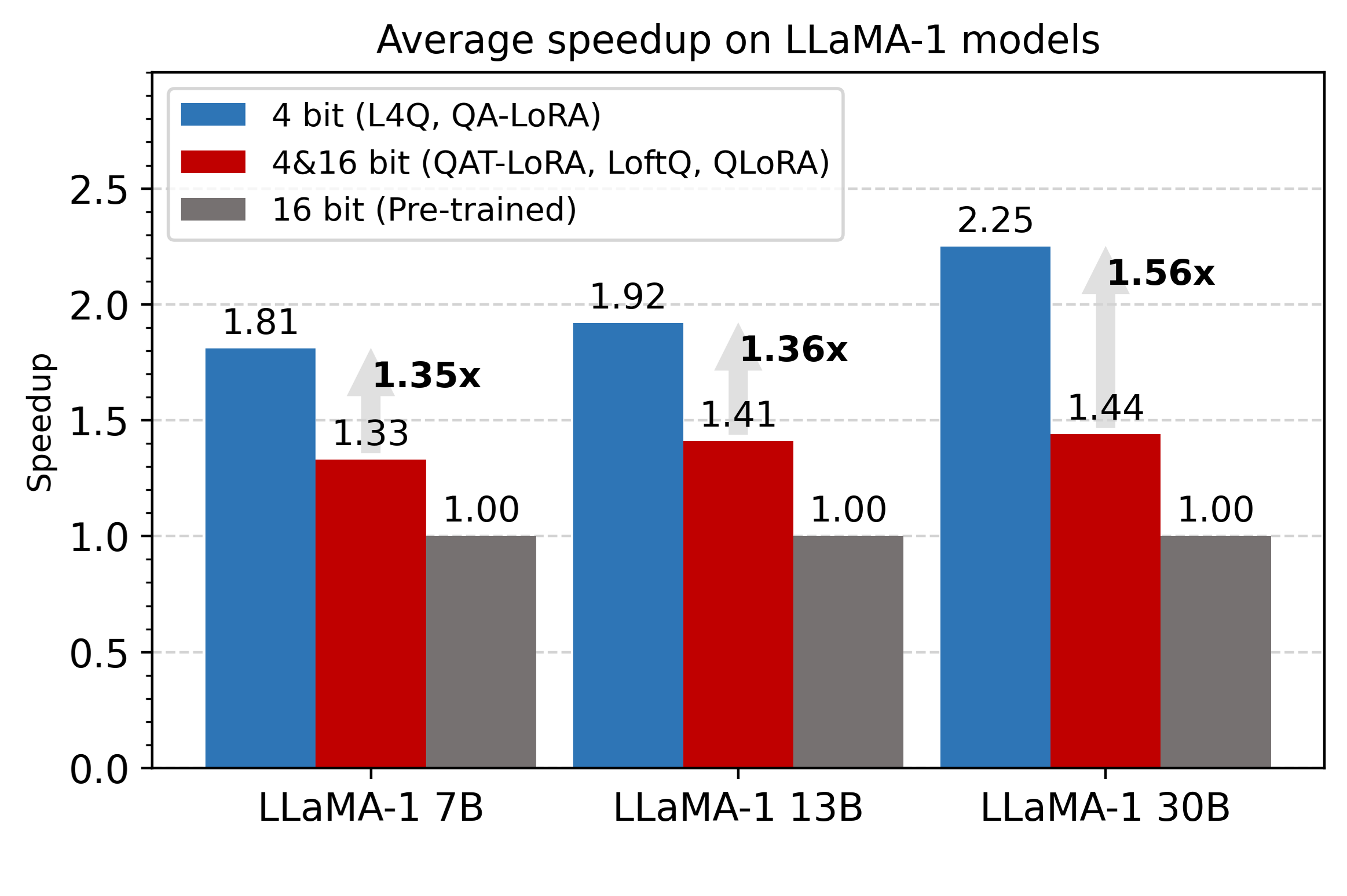}
   \vspace{-2mm}
    \caption{
    The average inference speedup of quantized models compared to pre-trained models.
    }
    \label{fig:speedup}
\vspace{-2mm}
\end{figure}

\begin{table*}[!ht]
    \centering
    \caption{Accuracy (\%) evaluation results with 4-bit quantization. We present the bit precision under each methods. The notation ‘4\&16’ refers to the use of 16-bit LoRA parameters with the quantized weights for inference.}
    \vspace{-4pt}
    \renewcommand{\arraystretch}{0.9}
    \label{table:csqa_mmlu_4bit}
    \resizebox{\textwidth}{!}{
    \begin{tabular}{l l
    |P{1.6cm}P{1.6cm}|P{1.6cm}P{1.6cm}|P{1.6cm}P{1.6cm}P{1.6cm}|P{1.6cm}P{1.6cm}}
    \toprule
    \rowcolor{Gray}
             ~ & ~ &  \small \bf Pre-trained & \small \bf LoRA &    \small \bf GPTQ &    \small \bf OmniQ &   \small  \bf LoftQ\textsuperscript{*} &   \small  \bf QLoRA\textsuperscript{*} &   \small  \bf QA-LoRA &  \small   \bf QAT-LoRA &   \small  \bf L4Q \\
    \rowcolor{Gray}
        Model & \bf Benchmark & 16 & 16 & 4 & 4 & 4\&16 & 4\&16 & 4 & 4\&16 & 4 \\
        \midrule
        OpenLLaMA 3B & CSQA & 54.8 & 55.9 & 50.7 & 54.1 & 54.2 & 54.4 & 54.5 & 54.6 & \bf 55.0 \\
        \midrule
        LLaMA-3 8B & CSQA & 65.6 & 67.2 & 57.9 & 64.9 & 63.3 & 58.6 & 58.0 & 65.7 & \bf 66.8 \\
        \midrule
        LLaMA-1 7B & CSQA & 61.7 & 63.4 & 59.4 & 58.1 & 62.6 & 61.3 & 61.3 & 62.4 & \bf 62.7 \\
        ~ & MMLU\tiny 0-shot & 32.5 & 36.3 & 28.3 & 30.9 & 33.0 & 32.8 & 34.5 & 33.8 & \bf 34.9 \\
        ~ & MMLU\tiny 5-shot & 35.1 & 36.7 & 32.7 & 33.3 & 35.1 & 33.6 & 35.6 & 34.8 & \bf 35.7 \\
        \midrule
        LLaMA-1 13B & CSQA & 63.8 & 65.2 & 63.5 & 60.4 & 64.2 & 63.8 & 62.5 & 64.4 & \bf 64.5 \\
        ~ & MMLU\tiny 0-shot & 43.6 & 44.3 & 40.1 & 42.6 & 42.4 & 42.1 & 42.4 & 42.0 & \bf 43.2 \\
        ~ & MMLU\tiny 5-shot & 46.3 & 47.0 & 45.7 & 45.7 & 45.4 & 45.9 & 45.8 & 45.5 & \bf 46.0 \\
        \midrule
        LLaMA-1 33B & CSQA & 67.4 & 68.3 & 65.7 & 62.9 & 67.4 & 66.2 & 65.3 & 67.3 & \bf 67.5 \\
        ~ & MMLU\tiny 0-shot & 53.0 & 54.4 & 51.4 & 52.0 & 51.8 & 51.0 & 48.9 & 52.3 & \bf 53.3 \\
        ~ & MMLU\tiny 5-shot & 56.4 & 57.6 & 55.7 & 55.8 & 56.4 & 55.6 & 55.0 & 56.7 & \bf 56.7 \\
        \midrule
        LLaMA-2 7B & CSQA & 61.9 & 63.3 & 60.7 & 59.5 & 61.7 & 61.3 & 61.0 & 61.9 & \bf 63.6 \\
        ~ & MMLU\tiny 0-shot & 41.6 & 43.9 & 37.1 & \bf 41.0 & 38.5 & 38.6 & 38.9 & 37.9 & 40.9 \\
        ~ & MMLU\tiny 5-shot & 45.4 & 46.0 & 42.9 & 45.4 & 43.7 & 44.6 & 44.4 & 43.8 & \bf 45.5 \\
        \midrule
        LLaMA-2 13B & CSQA & 65.0 & 66.5 & 64.4 & 59.9 & 64.9 & 64.0 & 64.5 & 64.7 & \bf 65.8 \\
        ~ & MMLU\tiny 0-shot & 52.1 & 52.5 & 50.0 & 51.8 & 51.7 & 50.7 & 50.4 & 50.7 & \bf 51.9 \\
        ~ & MMLU\tiny 5-shot & 54.8 & 55.7 & 54.7 & 54.7 & 54.5 & 54.2 & 54.1 & 53.8 & \bf 55.2 \\
        \midrule
        Mistral-v0.1 7B & CSQA & 66.2 & 66.4 & 65.3 & 64.7 & 60.7 & 65.8 & 65.4 & 64.5 & \bf 66.1 \\
        ~ & MMLU\tiny 0-shot & 60.2 & 60.6 & 57.6 & 58.4 & 45.2 & 58.7 & 56.5 & 58.8 & \bf 59.0 \\
        ~ & MMLU\tiny 5-shot & 62.6 & 62.9 & 61.0 & 61.0 & 45.7 & 61.1 & 61.2 & 60.2 & \bf 61.4 \\
        \bottomrule
    \end{tabular}
    }
\end{table*}

\begin{table*}[!ht]
    \centering
    \caption{Accuracy (\%) evaluation results with 3-bit quantization. We present the bit precision under each methods. The notation ‘3\&16’ refers to the use of 16-bit LoRA parameters with the quantized weights for inference.}
    \vspace{-4pt}
    \renewcommand{\arraystretch}{0.9}
    \label{table:csqa_mmlu_3bit}
    \resizebox{\textwidth}{!}{
        \begin{tabular}{l l
    |P{1.6cm}P{1.6cm}|P{1.6cm}P{1.6cm}|P{1.6cm}P{1.6cm}P{1.6cm}|P{1.6cm}P{1.6cm}}
    \toprule
    \rowcolor{Gray}
             ~ & ~ &  \small \bf Pre-trained & \small \bf LoRA &    \small \bf GPTQ &    \small \bf OmniQ &   \small  \bf LoftQ\textsuperscript{*} &   \small  \bf QLoRA\textsuperscript{*} &   \small  \bf QA-LoRA &  \small   \bf QAT-LoRA &   \small  \bf L4Q \\
    \rowcolor{Gray}
        Model & \bf Benchmark & 16 & 16 & 3 & 3 & 3\&16 & 3\&16 & 3 & 3\&16 & 3 \\
    \midrule
    OpenLLaMA 3B & CSQA & 54.8 & 55.9 & 52.2 & 50.0 & 38.1 & 51.0 & 51.5 & 53.2 & \bf 54.0 \\
    \midrule
    LLaMA-3 8B & CSQA & 65.6 & 67.2 & 53.5 & 58.7 & 48.6 & 55.7 & 56.6 & 63.1 & \bf 63.5 \\
    \midrule
    LLaMA-1 7B & CSQA & 61.7 & 63.4 & 53.4 & 56.5 & 49.8 & 59.1 & 58.7 & 60.7 & \bf 61.2 \\
    ~ & MMLU\tiny 0-shot & 32.5 & 36.3 & 23.7 & 29.0 & 23.4 & 27.7 & 28.0 & 30.6 & \bf 30.6 \\
    ~ & MMLU\tiny 5-shot & 35.1 & 36.7 & 27.3 & 31.6 & 23.1 & 31.5 & 29.1 & 31.5 & \bf 31.8 \\
    \midrule
    LLaMA-1 13B & CSQA & 63.8 & 65.2 & 61.0 & 58.9 & 54.0 & 61.3 & 61.1 & 63.2 & \bf 63.4 \\
    ~ & MMLU\tiny 0-shot & 43.6 & 44.3 & 33.1 & 34.8 & 25.0 & 36.1 & 37.5 & 38.8 & \bf 40.7 \\
    ~ & MMLU\tiny 5-shot & 46.3 & 47.0 & 38.2 & 41.6 & 25.3 & 40.4 & 38.2 & 40.9 & \bf 41.8 \\
    \midrule
    LLaMA-1 33B & CSQA & 67.4 & 68.3 & 65.1 & 62.3 & 54.8 & 64.3 & 64.6 & 67.4 & \bf 67.4 \\
    ~ & MMLU\tiny 0-shot & 53.0 & 54.4 & 50.0 & 50.2 & 24.6 & 45.6 & 46.1 & 50.1 & \bf 50.5 \\
    ~ & MMLU\tiny 5-shot & 56.4 & 57.6 & 51.9 & 52.4 & 24.0 & 50.1 & 48.7 & 50.6 & \bf 53.1 \\
    \midrule
    LLaMA-2 7B & CSQA & 61.9 & 63.3 & 57.6 & 57.9 & 34.7 & 57.6 & 56.3 & 57.4 & \bf 61.3 \\
    ~ & MMLU\tiny 0-shot & 41.6 & 43.9 & 31.3 & 34.3 & 22.9 & 32.5 & 31.0 & 31.5 & \bf 34.9 \\
    ~ & MMLU\tiny 5-shot & 45.4 & 46.0 & 37.5 & 37.7 & 24.2 & 37.6 & 37.5 & 36.8 & \bf 38.0 \\
    \midrule
    LLaMA-2 13B & CSQA & 65.0 & 66.5 & 61.7 & 59.9 & 39.3 & 62.5 & 61.7 & 64.3 & \bf 65.1 \\
    ~ & MMLU\tiny 0-shot & 52.1 & 52.5 & 46.3 & 46.3 & 23.5 & 46.8 & 46.4 & 45.9 & \bf 47.1 \\
    ~ & MMLU\tiny 5-shot & 54.8 & 55.7 & 50.4 & 50.2 & 26.0 & \bf 50.6 & 49.9 & 48.9 & 50.0 \\
    \midrule
    Mistral-v0.1 7B & CSQA & 66.2 & 66.4 & 61.8 & 61.4 & 58.5 & 63.0 & 62.3 & 61.6 & \bf 63.1 \\
    ~ & MMLU\tiny 0-shot & 60.2 & 60.6 & 50.5 & 54.3 & 35.8 & 52.2 & 50.5 & 52.4 & \bf 54.5 \\
    ~ & MMLU\tiny 5-shot & 62.6 & 62.9 & 49.6 & 55.9 & 37.1 & 53.6 & 51.7 & 54.0 & \bf 56.2 \\
    \bottomrule
    \end{tabular}
    }
\end{table*}

\textbf{Memory Cost for Fine-Tuning} \enspace
We measure the peak memory usage during fine-tuning of 4-bit LLMs, including L4Q and QAT-based baselines, as shown in Table~\ref{table:memory}. While QAT and QAT-LoRA incur significantly higher memory costs compared to LoRA, L4Q's memory usage remains comparable to that of LoRA.
This analysis shows that L4Q effectively balances the advantages of QAT and the memory efficiency of LoRA.
Further analysis of the training efficiency of L4Q and the baseline methods can be found in Appendix~\ref{appen:train_efficiency}. 

\noindent
\textbf{Inference Speedup} \enspace
We measure the inference speed of 16-bit pre-trained models and quantized models using LLaMA-1 models.
The average speedup of quantized models compared to full-precision 16-bit models is reported in Figure~\ref{fig:speedup}.
The quantized models include fully-quantized 4-bit models (L4Q and QA-LoRA), which contain only quantized parameters, and mixed-precision 4\&16-bit models (LoftQ\textsuperscript{*}, QLoRA\textsuperscript{*}, and QAT-LoRA), which use additional 16-bit LoRA parameters. The inference speed was measured with input batch sizes ranging from 1 to 64.
The 4-bit models achieve a speedup of 1.8$\times$ to 2.3$\times$ over the pre-trained models. Moreover, these 4-bit models achieve a 1.4$\times$ to 1.6$\times$ speedup compared to mixed-precision models, which are also quantized versions of LLMs. This demonstrates that the full integration of QAT and LoRA in L4Q plays a crucial role in inference speedup. Further analysis and details on speedup can be found in Appendix~\ref{appen:speedup}.

\noindent
\textbf{Accuracy Results}
We compare the CSQA and MMLU benchmark accuracy of baselines and L4Q.
Table~\ref{table:csqa_mmlu_4bit} and Table~\ref{table:csqa_mmlu_3bit} present a comprehensive comparison between baselines and the proposed L4Q.
Since previous quantization-aware PEFT methods involve a fine-tuning stage after PTQ, they generally achieve higher accuracy compared to PTQ methods.
L4Q further enhances accuracy by incorporating the QAT strategy, achieving highest accuracy compared to the baselines, and attaining 4-bit model accuracy comparable to that of 16-bit models. Moreover, L4Q consistently outperforms QAT-LoRA that keeps quantization and LoRA parameters decoupled. This highlights the advantage of L4Q in accuracy through the joint optimization of quantization and LoRA parameters.
This impact is more pronounced in 3-bit quantization, as some PTQ-based PEFT approaches experience significant accuracy degradation.
The detailed results are presented in Appendix~\ref{appen:l4q_csqa_mmlu}.

\section{Conclusion}
\label{sec:conclusion}
In this work, we introduce L4Q, a parameter-efficient quantization-aware fine-tuning method for large language models. L4Q enables element-wise adaptation of model weights for downstream tasks while simultaneously optimizing quantization parameters. This concurrent optimization ensures that the adaptation parameters effectively account for quantization errors.
We demonstrate the efficiency of L4Q, which significantly reduces training resource requirements compared to traditional QAT. Moreover, since the L4Q layer is designed to produce fully quantized low-bit model weights, it maintains inference efficiency, unlike QLoRA, LoftQ, or QAT-LoRA, which results in mixed precision models.
The effectiveness of L4Q as a QAT framework is further supported by experimental results in various task evaluations. L4Q consistently achieves superior quality maintenance in language tasks, demonstrating its enhanced adaptability compared to the QAT-LoRA and PTQ-based PEFT methods.

\section*{Limitations}
Our work focuses on efficient weight quantization methods for large language models (LLMs), but there are vertical approaches that could further enhance inference efficiency and effectiveness, integrated with our work.
Activation quantization offers a chance to further reduce computation costs when combined with weight quantization.
Similarly, KV cache compression could minimize memory overhead and latency, especially for long-context applications.
Finally, refinement of LoRA initialization schemes for quantized models may improve accuracy of the fine-tuned models.
We believe that integrating these approaches with L4Q could further improve LLM inference efficiency and effectiveness, which we leave to future work.

\section*{Ethical Considerations}
While our research contributes to the development and application of machine learning, particularly in language models, we recognize the potential societal implications associated with this work.
Based on the claims of the referenced sources — including datasets, code, and models — we believe that the artifacts do not violate personal identification rights or contain offensive content. All datasets, code, and models cited in this paper are publicly accessible and processed solely for research purposes. Our use of these artifacts is consistent with their intended purpose.

\section*{Acknowledgments}
This work was supported in part by Institute of Information \& communications Technology Planning \& Evaluation (IITP) grant funded by the Korea government (MSIT) (No.RS-2025-02273157: Development of Low Power Training/Inference Technologies based on AI Semiconductors, RS-2024-00395134, DPU-Centric Datacenter Architecture for Next-Generation AI Devices, No. 2021-0-01343: Artificial Intelligence Graduate School Program (Seoul National University)), Samsung Research Funding Center under Project SRFC-TC1603-53, and BK21 FOUR program at Seoul National University. (Corresponding Author: Jae-Joon Kim).

\bibliography{custom}

\newpage
\onecolumn
\section*{Appendix}
\appendix
\section{Details on Quantization Scale Learning Procedure}
\subsection{Quantization parameter update on QAT}

\label{appen:qat}

From the conditions and notations in Equation~\ref{eq:quant}, Equations ~\ref{eq:quant_lsq_s} and ~\ref{eq:quant_lsq_b} are derived as follows. First, the derivative of $s$ is presented as follows.

\begin{align}
\label{eq:appen_lsq_s}
\frac{\partial W_{q}}{\partial s}
= \frac{\partial }{\partial s}(\tilde{w} \times s + b)
= s\frac{\partial }{\partial s}(\tilde{w}) + \tilde{w}
= s\frac{\partial }{\partial s}(\text{round}\cdot \text{clamp}(w)) + \tilde{w}
\end{align}

By applying the STE, the rounding function is considered as an identity function. Therefore, the rounding function, combined with a clamp function $\tilde{\text{r}} := \text{round}\cdot \text{clamp}$, and its derivative is induced as follows.
Note that $w =\frac{W-b}{s}$.

\begin{align}
\label{eq:appen_ste}
\tilde{\text{r}}(w) = \begin{cases}
Qn, & if \enspace w < Qn \\
w, & if \enspace Q_{N} \leq w \leq Q_{P} \\
Qp, & if \enspace w > Q_{P}
\end{cases}
\quad
\frac{\partial}{\partial w}\tilde{\text{r}}(w) = \begin{cases}
1, & if \enspace Q_{N} \leq w \leq Q_{P} \\
0, & otherwise
\end{cases}
\end{align}

By applying the chain rule, the derivation of term $\tilde{\text{r}}(w) = \tilde{\text{r}}((W-b)/s)$ is expressed as below.

\begin{align}
\frac{\partial }{\partial s}(\tilde{\text{r}}(w)) = \frac{\partial \tilde{\text{r}}}{\partial w} \frac{\partial w}{\partial s} = \frac{\partial \tilde{\text{r}}}{\partial w} \frac{\partial}{\partial s}  (\frac{W-b}{s}) = \frac{\partial \tilde{\text{r}}}{\partial w} (- \frac{W-b}{s^2})
\end{align}

Therefore, Equation~\ref{eq:appen_lsq_s} can be represented with a value $w$ and quantized value $\tilde{w}$ as follows.

\begin{align}
\frac{\partial W_{q}}{\partial s} = s\frac{\partial \tilde{\text{r}}}{\partial w} (- \frac{W-b}{s^2}) + \tilde{w} = \frac{\partial \tilde{\text{r}}}{\partial w} (- \frac{W-b}{s}) + \tilde{w} = \begin{cases}
Qn, & if \enspace w < Qn \\
-w + \tilde{w}, & if \enspace Q_{N} \leq w \leq Q_{P} \\
Qp, & if \enspace w > Q_{P}
\end{cases}
\end{align}

Secondly, with a similar context above, the derivative of $b$ is presented as follows.

\begin{align}
\label{eq:appen_lsq_b}
\frac{\partial W_{q}}{\partial b} & = \frac{\partial }{\partial b}(\tilde{w} \times s + b) = s\frac{\partial }{\partial b}(\tilde{\text{r}}(w)) + 1 = s\frac{\partial \tilde{\text{r}}}{\partial w} (\frac{\partial}{\partial b} (\frac{W-b}{s})) + 1 \\
& = \frac{\partial \tilde{\text{r}}}{\partial w}(-1) + 1
= \begin{cases}
0, & if \enspace Q_{N} \leq w \leq Q_{P} \\
1, & otherwise
\end{cases}
\end{align}

We also note that the gradient of $W_q$ is presented as follows.

\begin{equation}
    \label{eq:appen_wq}
    \frac{\partial L}{\partial W_q} = \frac{\partial L}{\partial Y} X^{\top}
\end{equation}

As a result, the updates on the quantization scale and bias are calculated as multiplication of Equation~\ref{eq:appen_wq} with Equation~\ref{eq:appen_lsq_s} and with Equation~\ref{eq:appen_lsq_b}, respectively. This update helps calibrate the quantization function, effectively reducing quantization errors.

\newpage
\subsection{Quantization parameter and LoRA Parameter update on L4Q}
\label{appen:l4q}

In L4Q, as described in Equation~\ref{eq:l4q_forward}, the quantized weight $W_q$ is obtained as follows.
First, the pre-trained model weight $W_0$ and LoRA parameters are integrated to $W_{comb}=W_{0} +\alpha B A $.
Next, the integrated weight is quantized by the quantization parameters $s, b$.

Here, the LoRA parameters A and B are independent of the quantization parameters, scale $s$ and bias $b$.
Therefore, the derivatives of $s, b$ follow the same process as in Equation~\ref{appen:qat}, but with the term $w, \tilde{w}$ defined as follows. Note that $W_q = \tilde{w} \times s + b$.
\begin{align}
    \label{eq:appen_l4q_w}
    w = \frac{W_0 + \alpha BA - b}{s}, \enspace \tilde{w} = \tilde{\text{r}} (w) \quad s.t. \quad \tilde{\text{r}} = \text{round}\cdot \text{clamp}
\end{align}
Seen from the L4Q layer that integrates the LoRA parameters and quantization parameters together, $A, B$ are now considered as variables of $W_q$.
Therefore, from the conditions in Equation~\ref{eq:l4q_forward}, the derivative of $A, B$ is presented as follows.
\begin{align}
\label{eq:appen_l4q_0}
\frac{\partial L}{\partial A} = \frac{\partial W_{q}}{\partial A}\frac{\partial L}{\partial W_{q}}, \quad
\frac{\partial L}{\partial B} = \frac{\partial L}{\partial W_{q}}\frac{\partial W_{q}}{\partial B}
\end{align}
The derivatives $\frac{\partial w}{\partial A}$ and $\frac{\partial w}{\partial B}$ are then can be computed by applying the chain rule with $w$, as follows:
\begin{align}
\label{eq:appen_l4q_1}
\frac{\partial W_q}{\partial A} = \frac{\partial w}{\partial A}\frac{\partial W_{q}}{\partial w}, \quad
\frac{\partial W_q}{\partial B} = \frac{\partial W_q}{\partial w}\frac{\partial w}{\partial B}
\end{align}
From Equation~\ref{eq:appen_l4q_w}, the terms $\frac{\partial w}{\partial A}$, $\frac{\partial w}{\partial A}$, and $\frac{\partial W}{\partial w}$ can be expressed as follows:
\begin{align}
\label{eq:appen_l4q_2}
\frac{\partial w}{\partial A} = \frac{\alpha B^{\top}}{s}, \quad
\frac{\partial w}{\partial B} = \frac{\alpha A^{\top}}{s}, \quad
\frac{\partial W}{\partial w} = \frac{\partial }{\partial w} (\tilde{\text{r}}(w)s + b) = s\frac{\partial \tilde{\text{r}}}{\partial w} 
\end{align}
Therefore, by substitution of Equation~\ref{eq:appen_l4q_2} and applying STE on $\frac{\partial \tilde{\text{r}}}{\partial w}$ from Equation~\ref{eq:appen_ste} on Equation ~\ref{eq:appen_l4q_1}, the equation is simplified by the crossed-out products between the terms. As a result, the partial derivatives presented in Equation~\ref{eq:l4q_backward} can be derived as follows.
\begin{align}
\label{eq:appen_a}
    \frac{\partial W_{q}}{\partial A}
    = \frac{\partial w}{\partial A}\frac{\partial W_{q}}{\partial w}
    = (\frac{\alpha B^{\top}}{s})(\frac{s\partial \tilde{\text{r}}}{\partial w}) = \begin{cases}
    \alpha B^{\top}, & if \enspace Q_{N} \leq w \leq Q_{P} \\
    0, & otherwise 
\end{cases}
\end{align}

\begin{align}
 \label{eq:appen_b}
    \frac{\partial W_{q}}{\partial B} & 
    = \frac{\partial W}{\partial w}\frac{\partial w}{\partial B}
    = (s\frac{\partial \tilde{\text{r}}}{\partial w})(\frac{\alpha A^{\top}}{s}) = \begin{cases}
    \alpha A^{\top}, & if \enspace Q_{N} \leq w \leq Q_{P} \\
    0, & otherwise 
\end{cases}
\end{align}
Finally, substitution of Equation~\ref{eq:appen_a} and ~\ref{eq:appen_b} to Equation~\ref{eq:appen_l4q_0} derives the Equation~\ref{eq:appen_l4q_a} and ~\ref{eq:appen_l4q_b}.

\begin{equation}
    \label{eq:appen_l4q_a}
        \frac{\partial L}{\partial A} = \begin{cases}
        \alpha B^{\top}(\frac{\partial L}{\partial Y} X^{\top}), & \quad \mathrm{if} \enspace Q_{N} \leq w \leq Q_{P} \\
        0, & \quad otherwise
        \end{cases}
\end{equation}

\begin{equation}
    \label{eq:appen_l4q_b}
        \frac{\partial L}{\partial B} = \begin{cases}
        \alpha (\frac{\partial L}{\partial Y}X^{\top})A^{\top}, & \quad \mathrm{if} \enspace Q_{N} \leq w \leq Q_{P} \\
        0,  & \quad otherwise
        \end{cases}
\end{equation}

This form closely resembles the original backpropagation structure of the LoRA parameters $A, B$ as shown in Equation~\ref{eq:lora_backward}, where the updates are expressed as $\frac{\partial L}{\partial A} = \alpha \frac{\partial L}{\partial \tilde{X}} X^{\top} = \alpha (B^{\top}\frac{\partial L}{\partial Y}) X^{\top}$,
and $\frac{\partial L}{\partial B} = \alpha \frac{\partial L}{\partial Y} \tilde{X}^{\top} = \alpha \frac{\partial L}{\partial Y}(AX)^{\top}$, respectively. 
However, in L4Q, this process includes an added gating condition on the quantized weights, which accounts for the integration of quantization into the LoRA parameters.
As a result, we conclude that the backward process of the L4Q layer, which integrates both quantization parameter learning and LoRA parameter adaptation, is designed to account for the impact of quantization on the LoRA parameter updates.

\newpage
\section{Quantization initialization}
\label{appen:quant_init}

We evaluate various quantization initialization schemes within L4Q, including method introduced in Section~\ref{sec:quantinit}, LSQ+~\citep{bhalgat2020lsq}, and conventional symmetric and asymmetric quantization parameter initialization. The methods are depicted as L4Q\textsubscript{init}, LSQ+\textsubscript{init}, Symm, Asymm, respectively. 
In specific, each methods can be represented as the equations below, with quantization scale $s$ and bias $b$ and group-wise aligned model weight $W$ with quantization bit-width $n$ and $Q_{N}=-2^{n-1}, Q_{P}=2^{n-1}-1$.

\begin{align}
    \label{eq:quant_inits}
        \text{LSQ+}_{\text{init}}: \quad & s = \frac{Max(|\mu - 3\sigma(W)|, |\mu + 3\sigma(W)|)}{2^{n-1}} \\
        & b = 0 \\
        \text{Symm}: \quad & s = \frac{Max(Abs(W))}{2^{n-1}} \\
        & b = 0 \\
        \text{Asymm}: \quad & s = \frac{Max(W) - Min(W)}{Q_{P} - Q_{N}} \\
        & b = Max(W) - s\times Q_{P} = Min(W) - s\times Q_{N} \\
        \text{L4Q}_{\text{init}}: \quad & s = Max(|\frac{Min(W)}{Qn}|, |\frac{Max(W)}{Qp}|) \\
        & b = 0
\end{align}

We report the detailed model accuracy evaluation results of L4Q fine-tuning across different initialization methods, along with the quantization error and clipping error for each method, measured both at the initialization point and the end of the training.The LLaMA-2 7B model was trained for 12,800 iterations with a batch size of 128, using the same hyperparameters as in the main evaluation.

As shown in Table~\ref{table:quant_init}, while the overall quantization error remains relatively consistent across initialization methods and fine-tuned states, the clipping error exhibits significant variation. The clipping error reflects the number of values clipped during quantization, and different initialization methods lead to varying degrees of clipping throughout training. Notably, L4Q achieves the lowest clipping error and the highest model accuracy, demonstrating the effectiveness of its initialization strategy.

\begin{table}[!ht]
    \caption{MMLU 5-shot benchmark and the sum of quantization errors for various quantization parameter initialization methods within L4Q on the LLaMA-2 7B model. Quantization errors are represented in order of $10^6$ and clipping errors are represented in order of $10^3$.}
    \label{table:quant_init}
    \centering
    \resizebox{\textwidth}{!}{
    \begin{tabular}{l l l |ccccc| cc| cc}
    \toprule
     \rowcolor{Gray}
     ~ & ~ & ~ & \multicolumn{5}{c|}{\bf MMLU 5-shot} & \multicolumn{2}{c|}{\bf Initial} & \multicolumn{2}{c}{\bf Post-train} \\
     \rowcolor{Gray}
     \bf Model & \bf Method & \bf \#Bits & \bf Human. & \bf STEM & \bf Social. & \bf Others & \bf Average & \small E$_{quant}$  & \small E$_{clip}$ & \small E$_{quant}$ & \small E$_{clip}$\\
     \midrule
     LLaMA-2 7B & LSQ+ & 4 & 26.7 & 26.8 & 26.2 & 22.9 & 25.7 & 11.8 & 278.0 & 11.8 & 360.6 \\
     ~ & Symm & 4 & 40.8 & 35.9 & 48.2 & 50.1 & 43.5 & 11.1 & 260.0 & 11.0 & 282.1 \\
     ~ & Asymm & 4 & 41.0 & 37.1 & 49.7 & 50.2 & 44.2 & 10.5 & 0.0 & 10.5 & 64.7 \\
     ~ & L4Q & 4 & 42.9 & 37.7 & 50.5 & 51.9 & 45.3 & 11.4 & 0.0 & 11.6 & 36.1 \\
     \arrayrulecolor{black}
    \bottomrule
    \end{tabular}
    }
\end{table}

\newpage
\section{Ablative Study on L4Q Hyperparameters}
\label{appen:l4q_hyperparam}
\subsection{LoRA Rank Size}
\label{appen:lora_r}

We investigated the effect of LoRA rank size on L4Q training. Using the LLaMA-2 7B model, we conducted training over 12,800 iterations with 128 batches. The remaining training conditions are consistent with the main experiments.
The evaluation results for CSQA and MMLU are presented in Table~\ref{table:lora_rank_csqa} and Table~\ref{table:lora_rank_mmlu}, respectively. 

\begin{table}[!ht]
    \centering
    \caption{Commonsense QA benchmark result on LLaMA-2 7B model. The numbers represent accuracy (\%) for each task.}
    \resizebox{\textwidth}{!}{
    \begin{tabular}{ll | c c cccccc}
        \toprule
        \rowcolor{Gray} \bf Model & \bf Rank & \bf HellaSwag & \bf PIQA & \bf ARC-c & \bf ARC-e & \bf Winogrande & \bf BoolQ & \bf OBQA & \bf Average \\
        \midrule
        LLaMA-2 7B & 1 & 57.6 & 78.4 & 45.5 & 77.0 & 68.2 & 77.4 & 34.0 & 62.6 \\ 
        ~ & 2 & 57.4 & 78.5 & 45.1 & 76.2 & 69.3 & 77.6 & 34.0 & 62.6 \\ 
        ~ & 4 & 57.5 & 78.2 & 46.1 & 77.1 & 68.7 & 78.1 & 35.4 & 63.0 \\ 
        ~ & 8 & 56.9 & 78.5 & 46.3 & 78.1 & 69.3 & 77.8 & 34.8 & 63.1 \\ 
        ~ & 16 & 57.2 & 78.1 & 46.3 & 77.2 & 68.7 & 78.9 & 33.4 & 62.8 \\ 
        ~ & 32 & 57.8 & 78.4 & 46.2 & 77.1 & 68.7 & 78.2 & 35.8 & 63.2 \\ 
        ~ & 64 & 57.4 & 78.6 & 46.1 & 77.1 & 69.5 & 78.5 & 34.8 & 63.1 \\ 
        ~ & 128 & 57.5 & 78.1 & 46.0 & 77.0 & 68.8 & 78.4 & 35.4 & 63.0 \\ 
        \bottomrule
    \end{tabular}
    }
    \label{table:lora_rank_csqa}
\end{table}

\begin{table}[!ht]
    \centering
    \caption{MMLU benchmark result on LLaMA-2 7B model. The numbers represent accuracy (\%) for each category.}
    \resizebox{\textwidth}{!}{
    \begin{tabular}{ll | ccccc ccccc}
        \toprule
        \rowcolor{Gray}
        ~ & ~ & \multicolumn{5}{c|}{\bf 0-shot} & \multicolumn{5}{c}{\bf 5-shot} \\
        \rowcolor{Gray}
        \bf Model & \bf Rank & \bf Hums.	& \bf STEM	& \bf Social & \bf Others & \bf Avg. & \bf Hums. & \bf STEM & \bf Social & \bf Others	& \bf Avg. \\
        
        \midrule
        LLaMA-2 7B & 1 & 37.4 & 33.3 & 43.4 & 44.0 & 39.4 & 40.8 & 36.8 & 48.2 & 49.2 & 43.5 \\ 
        ~ & 2 & 38.1 & 31.6 & 41.6 & 42.2 & 38.4 & 42.2 & 35.2 & 48.6 & 49.0 & 43.7 \\ 
        ~ & 4 & 36.3 & 33.4 & 42.7 & 43.2 & 38.7 & 42.5 & 36.6 & 50.2 & 51.2 & 44.9 \\ 
        ~ & 8 & 39.5 & 34.6 & 45.0 & 45.0 & 41.0 & 42.7 & 36.7 & 50.3 & 51.7 & 45.0 \\ 
        ~ & 16 & 38.7 & 35.8 & 45.7 & 45.8 & 41.3 & 42.0 & 36.6 & 49.4 & 49.7 & 44.3 \\ 
        ~ & 32 & 39.4 & 35.0 & 46.1 & 45.6 & 41.3 & 42.4 & 37.1 & 49.8 & 49.0 & 44.4 \\ 
        ~ & 64 & 39.6 & 35.0 & 45.8 & 47.2 & 41.7 & 43.6 & 37.4 & 50.9 & 50.9 & 45.6 \\ 
        ~ & 128 & 38.8 & 35.4 & 44.8 & 44.7 & 40.7 & 42.8 & 36.6 & 50.3 & 50.6 & 44.9 \\ 
        \bottomrule
    \end{tabular}
    }
    \label{table:lora_rank_mmlu}
\end{table}

Increasing the rank beyond 4 does not lead to significant performance improvements, which aligns with the observations in the original LoRA paper~\citep{hu2021lora}. Therefore, we generally applied a rank size of 4, considering that higher rank sizes introduce memory and computational overhead during training.

\newpage
\subsection{Quantization Group Size}
\label{appen:quant_group}

We investigated the effect of quantization group size on L4Q training. Using the LLaMA-2 7B model, we conducted experiments with group size 32 to 128 with the same training conditions in the main experiments.
The evaluation results for Commonsense QA and MMLU are presented in Table~\ref{table:quant_group_csqa} and Table~\ref{table:quant_group_mmlu}, respectively. 

\begin{table}[!ht]
    \centering
    \caption{Commonsense QA benchmark result on LLaMA-2 7B model. The numbers represent accuracy (\%) for each task.}
    \resizebox{\textwidth}{!}{
    \begin{tabular}{ll | c c c c c c c c}
        \toprule
        \rowcolor{Gray} \bf Model & \bf Group Size & \bf Hellaswag & \bf PIQA & \bf ARC-c & \bf ARC-e & \bf Winogrande & \bf BoolQ & \bf OBQA & \bf Average \\
        \midrule
        LLaMA-2 7B & 128 & 57.2 & 78.8 & 47.1 & 76.9 & 70.2 & 80.4 & 34.8 & 63.6 \\ 
        ~ & 64  & 57.5 & 77.5 & 46.7 & 78.3 & 70.2 & 80.7 & 34.8 & 63.7 \\ 
        ~ & 32  & 57.6 & 77.6 & 47.6 & 78.2 & 70.1 & 80.6 & 35.4 & 63.9 \\ 
        \bottomrule
    \end{tabular}
    }
    \label{table:quant_group_csqa}
\end{table}

\begin{table}[!ht]
    \centering
    \caption{MMLU benchmark result on LLaMA-2 7B model. The numbers represent accuracy (\%) for each category.}
    \resizebox{\textwidth}{!}{
    \begin{tabular}{ll | ccccc ccccc}
        \toprule
        \rowcolor{Gray}
        ~ & ~ & \multicolumn{5}{c|}{\bf 0-shot} & \multicolumn{5}{c}{\bf 5-shot} \\
        \rowcolor{Gray}
        \bf Model & \bf Group Size & \bf Hums. & \bf STEM & \bf Social & \bf Others & \bf Avg. & \bf Hums. & \bf STEM & \bf Social & \bf Others & \bf Avg. \\
        \midrule
        LLaMA-2 7B & 128 & 38.7 & 33.8 & 45.6 & 46.4 & 40.9 & 42.9 & 37.7 & 50.5 & 51.9 & 45.5 \\ 
        ~ & 64  & 38.2 & 35.6 & 47.2 & 46.5 & 41.5 & 43.8 & 37.0 & 52.2 & 52.7 & 46.3 \\ 
        ~ & 32  & 39.5 & 35.9 & 47.8 & 46.4 & 42.1 & 44.4 & 38.2 & 52.3 & 52.5 & 46.7 \\ 
        \bottomrule
    \end{tabular}
    }
    \label{table:quant_group_mmlu}
\end{table}

Having a fine-grained quantization group size leads to performance improvements, which aligns with the observations in the conventional group-wise quantization works~\citep{frantar2023gptq}.
We applied a quantization with the group size of 128 considering that smaller quantization group sizes introduce memory and computational overhead during inference and training.

\newpage
\section{Experimental Settings}
\label{appen:ex_settings}
The baselines and L4Q are trained with AdamW optimizer~\cite{loshchilov2019decoupled} with a weight decay of 0.01. For the learning rate scheduler, a cosine decay scheduler with a linear warm-up through 10\% of the total training steps. Learning rates are presented in Table~\ref{table:ex_settings}.

\begin{table}[!ht]
\centering
\caption{Learning rate conditions used to fine-tuning on each models for L4Q and baselines: QLoRA\textsuperscript{*}, QA-LoRA, and QAT-LoRA.}
\label{table:ex_settings}
\renewcommand{\arraystretch}{1.2}
\begin{tabular}{l cccc}
\toprule
 \rowcolor{Gray}~ & \multicolumn{4}{c}{\bf Methods}       \\
 \rowcolor{Gray} \bf Model & \multicolumn{1}{c}{\bf QLoRA\textsuperscript{*}} & \bf QA-LoRA & \bf QAT-LoRA & \bf L4Q \\
\midrule
OpenLLaMA 3B & $1\times 10^{-5}$ & $2\times 10^{-5}$ & $5\times 10^{-5}$ & $5\times 10^{-5}$ \\
LLaMA-1 7B & $1\times 10^{-5}$ & $2\times 10^{-5}$ & $5\times 10^{-5}$ & $5\times 10^{-5}$ \\
LLaMA-1 13B & $1\times 10^{-5}$ & $5\times 10^{-5}$ & $4\times 10^{-5}$ & $4\times 10^{-5}$ \\
LLaMA-1 33B & $1\times 10^{-5}$ & $5\times 10^{-5}$ & $2\times 10^{-4}$ & $2\times 10^{-4}$ \\
LLaMA-2 7B & $2\times 10^{-5}$ & $2\times 10^{-5}$ & $2\times 10^{-4}$ & $2\times 10^{-4}$ \\
LLaMA-2 13B & $2\times 10^{-5}$ & $2\times 10^{-5}$ & $2\times 10^{-4}$ & $2\times 10^{-4}$ \\
Mistral-v0.1 7B & $1\times 10^{-5}$ & $5\times 10^{-6}$ & - & $5\times 10^{-6}$ \\
        \arrayrulecolor{black}
        \bottomrule
\end{tabular}
\end{table}

\begin{figure}[!ht]
    \centering
    \includegraphics[width=0.6\linewidth]{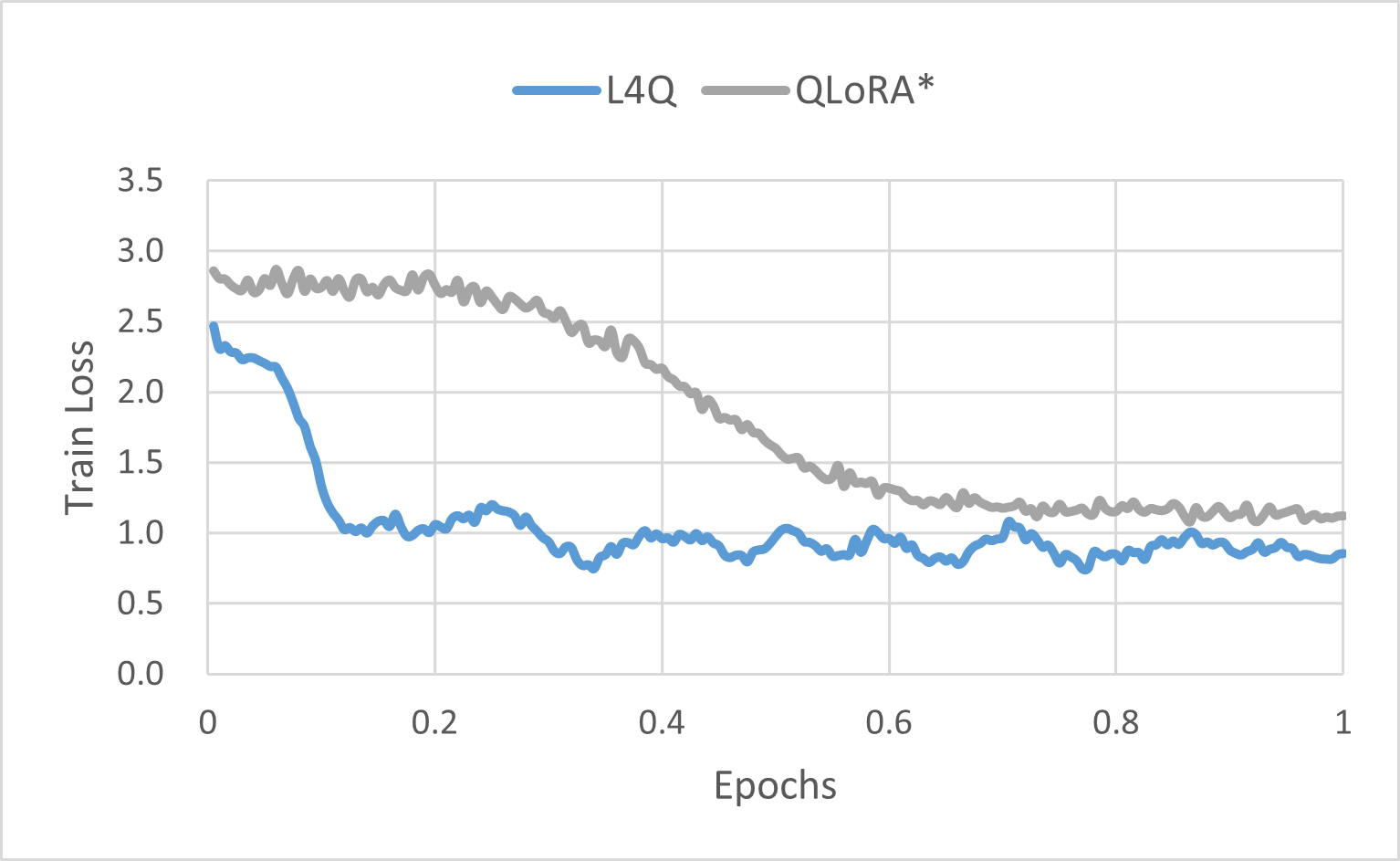}
    \caption{Train loss curve of L4Q and QLoRA. With a same training condition, L4Q converges faster than QLoRA.} 
    \label{fig:train_loss}
\end{figure}

The batch size is set to 128. For baselines that utilize PTQ-based schemes, such as QLoRA\textsuperscript{*} and QA-LoRA, training is conducted for 50K iterations. For QAT-based methods, such as QAT, QAT-LoRA, and L4Q, training is conducted for 25K iterations. This reduction in training length for QAT-based methods is due to their faster convergence, as illustrated in Figure~\ref{fig:train_loss} with an example of LLaMA-2 7B.

Using the same training hyperparameters, including a learning rate of $2\times 10^{-5}$, the joint training of quantization parameters and LoRA weight parameters enables L4Q to converge more quickly. This allows for halving the training length, which also helps mitigate overfitting.

Additionally, the training sequence length is set to match or exceed the maximum sequence length of the dataset, which is 2048. The only exception is the 33B model with L4Q, where the training sequence length is set to 128.

\newpage
\section{Training Efficiency}
\label{appen:train_efficiency}
\subsection{Memory Usage}
\label{appen:memory}

We report the memory usage of QAT (LSQ), quantization-aware PEFT baselines (LoftQ, QLoRA, QA-LoRA, QAT-LoRA), and L4Q in Table~\ref{table:appen_memory}. Note that QLoRA reduces memory usage further by employing paged optimizers, a technique that can also be applied to other fine-tuning methods, including L4Q. We plan to explore this vertical implementation in future work.
As quantization-aware PEFT methods utilize pre-quantized model weights, the memory usage of L4Q and quantization-aware PEFT methods differs by the amount of the reduced memory usage of model weights during inference. 

\begin{table}[!ht]
    \centering
    \begin{tabular}{lP{1.6cm}P{1.6cm}P{1.6cm}P{1.6cm}}
    \toprule
     \rowcolor{Gray}& \small \bf OpenLLaMA & \multicolumn{3}{c}{\small \bf LLaMA} \\
     \rowcolor{Gray}\bf Methods & \bf 3B & \bf 7B & \bf 13B & \bf 33B \\
     \arrayrulecolor{black}\midrule
     LoRA  & 15.1 & 25.1 & 43.8 & 71.9 \\
     \arrayrulecolor{lightgray}\cmidrule{1-5}
     LoftQ  & 5.2 & 7.9 & 19.6 & 31.9 \\
     QLoRA  & 5.2 & 7.9 & 19.6 & 31.9 \\
     QA-LoRA  & 7.8 & 14.8 & 27.8 & 67.2 \\
     \arrayrulecolor{lightgray}\cmidrule{1-5}
     LSQ  & 44.2 & 79.5  & OOM & OOM \\
     QAT-LoRA  & 22.6 & 41.9 & 70.6 & OOM \\
     L4Q  & 15.3 & 25.4 & 44.3 & 73.2 \\
    \arrayrulecolor{black}\bottomrule
    \end{tabular}
    \caption{Memory cost (GB) for fine-tuning LLMs on a single NVIDIA A100 GPU. (OOM: Out of Memory)}
    \label{table:appen_memory}
\end{table}

\subsection{Training Time}
\label{appen:time}

We report the total training time of L4Q, QAT (LSQ), and the quantization-aware PEFT baselines(QLoRA\textsuperscript{*} and QA-LoRA) in Table~\ref{table:time}. While L4Q has a longer training time per step compared to the baselines due to gradient recomputation during the backpropagation stage, the reduced number of training steps enables L4Q to achieve similar overall training time performance. The training time for QAT on 13B and 33B models was measured using 2 and 4 NVIDIA A100 GPUs on a single node.

\begin{table}[!ht]
    \centering
    \caption{Training time (in hours) spent on fine-tuning on OpenLLaMA and LLaMA-1 models with a A100 GPU.}
    \begin{tabular}{lP{1.6cm}P{1.6cm}P{1.6cm}P{1.6cm}}
        \toprule
        \rowcolor{Gray} & \bf \small OpenLLaMA & \multicolumn{3}{c}{\bf \small LLaMA} \\
        \rowcolor{Gray} \bf Methods & \bf 3B & \bf 7B & \bf 13B & \bf 33B \\
        \arrayrulecolor{black}\midrule
        LSQ     & 8.6  & 17.1 & 35.4  & 76.9 \\
        \arrayrulecolor{lightgray}\cmidrule{1-5}
        QLoRA\textsuperscript{*}  & 4.5  & 9.9  & 18.0  & 38.4 \\
        QA-LoRA  & 5.0  & 11.2 & 19.8  & 39.6 \\
        L4Q     & 4.4  & 10.1 & 16.9  & 37.9 \\
        \arrayrulecolor{black}\bottomrule
    \end{tabular}
    \label{table:time}
\end{table}

\newpage
\section{Throughput and Speedup of fully-quantized models and mixed-precision models}
\label{appen:speedup}

We investigate the throughput and speedup of fully-quantized models and mixed-precision models, demonstrating that, although the number of LoRA parameters is negligible, it causes a noticeable drop in throughput when its forward path is not merged with that of the base linear layer.
Using the LLaMA-1 7B, 13B, and 33B models, we conducted experiments to measure throughput (tokens per second) and compare speedup.
In both fully-quantized and mixed-precision models, uniform quantization is applied to the linear layers, except for the head layer (lm\_head), using the EXLLaMA2 kernel\footnote{https://github.com/turboderp/exllamav2.git}, which is designed for 4-bit weight-only quantized inference.
For fp16 computations in LoRA within mixed-precision models or the baseline, default GEMM kernels are used.
We measured the elapsed time for inferencing 512 tokens over 2000 data points with batch sizes ranging from 1 to 64, calculating throughput by dividing the number of tokens, which is set to be 512, by the elapsed time. The results are presented in Table~\ref{table:speedup}.

\begin{table}[!ht]
    \centering
    \caption{Throughput (tokens/sec) and Speedup for LLaMA models. L4Q represents fully-quantized models, and QLoRA\textsuperscript{*} represents mixed-precision models. 'OOM' indicates out-of-memory cases.}
    \resizebox{\textwidth}{!}{
    \begin{tabular}{l l  ccccccc  c}
    \toprule
        \rowcolor{Gray} ~ & ~ & \multicolumn{7}{c}{\bf Batch size} & ~ \\

        \rowcolor{Gray}\bf Model & \bf Method  & \bf 1 & \bf 2 & \bf 4 & \bf 8 & \bf 16 & \bf 32 & \bf 64 & \bf Speedup \\
        \arrayrulecolor{black}
        \midrule
        LLaMA-1 7B & L4Q & 38.04 & 75.12 & 148.63 & 216.51 & 255.64 & 276.43 & 318.43 & 1.81 \\
        ~ & QLoRA\textsuperscript{*} & 24.80 & 47.88 & 96.51 & 184.06 & 234.79 & 247.79 & 299.94 & 1.33 \\
        ~ & None & 17.04 & 33.81 & 67.27 & 124.13 & 199.19 & 241.31 & OOM & 1.00 \\
        \arrayrulecolor{gray}
        \midrule
        LLaMA-1 13B & L4Q & 30.68 & 59.53 & 115.78 & 144.44 & 160.95 & 191.20 & OOM & 1.92 \\
        ~ & QLoRA\textsuperscript{*} & 19.71 & 38.90 & 77.83 & 128.67 & 150.72 & 156.16 & OOM & 1.41 \\
        ~ & None & 13.67 & 26.97 & 53.23 & 85.47 & 124.17 & OOM & OOM & 1.00 \\
        \arrayrulecolor{gray}
        \midrule
        LLaMA-1 30B & L4Q & 20.43 & 40.05 & 64.00 & 73.29 & 79.77 & OOM & OOM & 2.25 \\
        ~ & QLoRA\textsuperscript{*} & 13.22 & 25.48 & 50.81 & 66.89 & 75.11 & OOM & OOM & 1.44 \\
        ~ & None & 9.13 & 17.68 & OOM & OOM & OOM & OOM & OOM & 1.00 \\
        \arrayrulecolor{black}
        \bottomrule
    \end{tabular}
    }
    \label{table:speedup}
\end{table}

Fully-quantized models demonstrate a speedup of over 1.8x, while mixed-precision models achieve a maximum speedup of 1.4x, despite using the same quantization scheme and execution kernel, compared to the fp16 baselines. As a result, fully-quantized models achieve a 30\% to 50\% greater speedup compared to mixed-precision models.
This demonstrates that L4Q, which produces fully-quantized models, offers higher inference efficiency and better hardware utilization than conventional quantization-aware PEFT methods, such as QLoRA and LoftQ, which retain unmerged forward paths for LoRA.

\newpage
\section{Detailed Result on Main Evaluations}
\label{appen:l4q_csqa_mmlu}

We present the Commonsense QA and MMLU benchmark results with averaged accuracy score on Section~\ref{sec:experiments}. We present the detailed results of each benchmarks composed of several categories of tasks in Table~\ref{table:csqa_a} to Table~\ref{table:mmlu_b} below.
Through evaluation, we demonstrate that L4Q generally achieves higher accuracy in low-bit quantized models compared to both PTQ methods and PTQ-based fine-tuning methods. Notably, L4Q surpasses the pre-trained models on the Commonsense QA benchmarks and on the MMLU benchmarks with LLaMA-1 7B and 33B models. In contrast, PTQ-based fine-tuning methods, including those that incorporate high-precision LoRA weights, show lower performance compared to both L4Q and the pre-trained models. These results emphasize the challenges of recovering from quantization errors with PTQ alone and highlight the effectiveness of L4Q's joint quantization and fine-tuning scheme.

\renewcommand{\thetable}{13a}
\begin{table}[!ht]
    \centering
    \caption{Commonsense QA benchmark result. The numbers represent accuracy (\%) of each task.}
    \renewcommand{\arraystretch}{0.1}
    \resizebox{0.9\textwidth}{!}{
    \begin{tabular}{l l l | ccccccc  c}
    \toprule
    \rowcolor{Gray}
        \bf Model & \bf Method & \bf \#Bits & \bf Hella. & \bf PIQA & \bf ARC-c & \bf ARC-e & \bf Winogr. & \bf BoolQ & \bf OBQA & \bf Avg. \\ 
        \arrayrulecolor{black}
        \midrule
        
        OpenLLaMA 3B & None & 16 & 48.8 & 75.0 & 33.9 & 69.2 & 61.6 & 66.9 & 28.2 & \bf 54.8 \\
        ~ & LoRA & 16 & 49.8 & 75.6 & 37.0 & 70.2 & 63.1 & 68.0 & 27.2 & \bf 55.9 \\
        \arrayrulecolor{lightgray}\cmidrule{2-11}
        ~ & GPTQ & 4 & 47.9 & 75.1 & 31.0 & 58.8 & 60.5 & 57.9 & 23.6 & \bf 50.7 \\
        ~ & OmniQ & 4 & 48.2 & 73.8 & 33.1 & 69.5 & 60.1 & 67.5 & 26.6 & \bf 54.1 \\
        ~ & LoftQ\textsuperscript{*} & 4\&16 & 48.0 & 74.5 & 34.0 & 68.6 & 60.9 & 67.2 & 26.0 & \bf 54.2 \\
        ~ & QLoRA\textsuperscript{*} & 4\&16 & 48.4 & 74.3 & 33.0 & 69.4 & 61.5 & 67.1 & 26.8 & \bf 54.4 \\
        ~ & QA-LoRA & 4 & 48.8 & 74.9 & 33.8 & 69.2 & 61.9 & 66.7 & 26.2 & \bf 54.5 \\
        ~ & QAT-LoRA & 4\&16 & 48.8 & 74.5 & 35.0 & 70.1 & 61.9 & 65.2 & 27.0 & \bf 54.6 \\
        ~ &   L4Q & 4 & 49.1 & 74.9 & 35.2 & 69.8 & 61.1 & 67.7 & 27.4 & \bf   55.0 \\
        \arrayrulecolor{lightgray}\cmidrule{2-11}
        ~ & GPTQ & 3 & 46.3 & 72.6 & 31.8 & 64.7 & 58.1 & 66.5 & 25.6 & \bf 52.2 \\
        ~ & OmniQ & 3 & 46.5 & 74.4 & 30.5 & 56.6 & 59.0 & 59.8 & 23.0 & \bf 50.0 \\
        ~ & LoftQ\textsuperscript{*} & 3\&16  & 27.9 & 57.3 & 19.5 & 37.3 & 51.0 & 61.9 & 12.0 & \bf 38.1 \\
        ~ & QLoRA\textsuperscript{*} & 3\&16 & 45.6 & 72.6 & 29.3 & 61.6 & 59.7 & 64.2 & 24.4 & \bf 51.0 \\
        ~ & QA-LoRA & 3 & 46.3 & 72.6 & 28.9 & 66.0 & 59.5 & 63.4 & 23.8 & \bf 51.5 \\
        ~ & QAT-LoRA & 3\&16 & 46.7 & 74.1 & 33.2 & 67.2 & 60.5 & 64.1 & 26.4 & \bf 53.2 \\
        ~ &   L4Q & 3 & 47.2 & 75.0 & 32.3 & 68.3 & 60.9 & 67.2 & 27.0 &  \bf  54.0 \\
        \arrayrulecolor{black}
        \midrule

        LLaMA-3 8B & None & 16 & 60.2 & 79.7 & 50.4 & 80.1 & 72.5 & 81.4 & 34.8 & \bf 65.6 \\ 
        ~ & LoRA & 16 & 60.6 & 79.9 & 53.8 & 82.7 & 74.7 & 82.8 & 36.0 & \bf 67.2 \\ 
        \arrayrulecolor{lightgray}
        \cmidrule{2-11}
        ~ & GPTQ & 4 & 56.8 & 69.3 & 31.8 & 62.6 & 73.1 & 78.5 & 33.0 & \bf 57.9 \\ 
        ~ & OmniQ & 4 & 59.2 & 78.6 & 49.6 & 79.7 & 72.5 & 81.1 & 33.4 & \bf 64.9 \\ 
        ~ & LoftQ\textsuperscript{*} & 4\&16 & 57.9 & 78.7 & 46.9 & 78.4 & 70.3 & 77.5 & 33.6 & \bf 63.3 \\ 
        ~ & QLoRA\textsuperscript{*} & 4\&16 & 56.7 & 69.5 & 33.3 & 66.3 & 72.9 & 78.4 & 33.0 & \bf 58.6 \\ 
        ~ & QA-LoRA & 4 & 56.3 & 69.3 & 32.3 & 65.7 & 72.3 & 78.6 & 31.6 & \bf 58.0 \\ 
        ~ & QAT-LoRA & 4\&16 & 59.1 & 79.8 & 50.1 & 80.2 & 74.0 & 83.0 & 34.2 & \bf 65.7 \\ 
        ~ & L4Q & 4 & 60.5 & 80.4 & 52.7 & 81.6 & 73.6 & 83.6 & 35.0 & \bf 66.8 \\ 
        \arrayrulecolor{lightgray}
        \cmidrule{2-11}
        ~ & GPTQ & 3 & 51.8 & 68.8 & 30.2 & 58.3 & 67.7 & 70.1 & 27.4 & \bf 53.5 \\ 
        ~ & OmniQ & 3 & 55.0 & 76.7 & 39.2 & 69.2 & 69.2 & 72.6 & 28.8 & \bf 58.7 \\ 
        ~ & LoftQ\textsuperscript{*} & 3\&16 & 35.9 & 68.8 & 29.8 & 57.7 & 59.0 & 67.9 & 20.8 & \bf 48.6 \\ 
        ~ & QLoRA\textsuperscript{*} & 3\&16 & 53.4 & 72.2 & 31.8 & 62.9 & 69.2 & 71.5 & 28.8 & \bf 55.7 \\ 
        ~ & QA-LoRA & 3 & 52.7 & 73.5 & 36.4 & 67.9 & 67.6 & 71.7 & 26.6 & \bf 56.6 \\ 
        ~ & QAT-LoRA & 3\&16 & 56.6 & 78.2 & 47.4 & 77.8 & 68.0 & 80.6 & 33.4 & \bf 63.1 \\ 
        ~ & L4Q & 3 & 56.5 & 78.1 & 47.8 & 78.6 & 69.2 & 82.0 & 32.2 & \bf 63.5 \\ 
        \arrayrulecolor{black}
        \midrule
        
        LLaMA-1 7B & None & 16 & 57.0 & 78.7 & 41.9 & 75.3 & 69.9 & 75.1 & 34.4 & \bf 61.7 \\
        ~ & LoRA & 16 & 58.3 & 78.8 & 45.7 & 76.1 & 70.6 & 78.7 & 35.4 & \bf 63.4 \\
        \arrayrulecolor{lightgray}\cmidrule{2-11}
        ~ & GPTQ & 4 & 53.9 & 77.7 & 40.3 & 73.5 & 67.9 & 72.9 & 30.0 & \bf 59.4 \\
        ~ & OmniQ & 4 & 55.7 & 77.7 & 38.8 & 67.5 & 65.3 & 72.5 & 29.2 & \bf 58.1 \\
        ~ & LoftQ\textsuperscript{*} & 4\&16 & 57.8 & 79.2 & 43.1 & 76.9 & 69.8 & 75.8 & 35.4 & \bf 62.6 \\
        ~ & QLoRA\textsuperscript{*} & 4\&16 & 56.7 & 78.9 & 41.8 & 75.2 & 70.0 & 74.6 & 32.2 & \bf 61.3 \\
        ~ & QA-LoRA & 4 & 57.2 & 78.9 & 41.2 & 74.9 & 70.6 & 73.6 & 32.6 & \bf 61.3 \\
        ~ & QAT-LoRA & 4\&16 & 57.7 & 78.9 & 44.7 & 75.3 & 68.9 & 75.8 & 35.6 & \bf 62.4 \\
        ~ &   L4Q & 4 & 57.8 & 79.1 & 45.3 & 76.0 & 69.5 & 76.1 & 34.8 &   \bf 62.7 \\
        \arrayrulecolor{lightgray}\cmidrule{2-11}
        ~ & GPTQ & 3 & 46.6 & 71.9 & 32.4 & 65.4 & 65.0 & 68.0 & 24.6 & \bf 53.4 \\
        ~ & OmniQ & 3 & 54.0 & 77.1 & 35.6 & 64.9 & 64.7 & 71.2 & 28.0 & \bf 56.5 \\
        ~ & LoftQ\textsuperscript{*} & 3\&16 & 43.4 & 68.9 & 33.0 & 65.5 & 56.5 & 58.5 & 23.0 & \bf 49.8 \\
        ~ & QLoRA\textsuperscript{*} & 3\&16 & 53.9 & 76.2 & 39.3 & 71.5 & 68.9 & 72.8 & 31.0 & \bf 59.1 \\
        ~ & QA-LoRA & 3 & 55.4 & 76.3 & 39.8 & 72.5 & 69.5 & 67.1 & 30.6 & \bf 58.7 \\
        ~ & QAT-LoRA & 3\&16 & 56.1 & 77.4 & 41.6 & 72.8 & 68.0 & 76.0 & 33.0 & \bf 60.7 \\
        ~ &   L4Q & 3 & 55.9 & 77.6 & 42.1 & 74.1 & 68.9 & 76.8 & 33.4 &   \bf 61.2 \\
        \arrayrulecolor{black}
        \bottomrule
    \end{tabular}
    }
    \label{table:csqa_a}
\end{table}

\renewcommand{\thetable}{13b}
\begin{table}[!ht]
    \centering
    \caption{Commonsense QA benchmark result. The numbers represent accuracy (\%) of each task.}
    \renewcommand{\arraystretch}{0.1}
    \resizebox{0.85\textwidth}{!}{
    \begin{tabular}{l l l | ccccccc  c}
    \toprule
    \rowcolor{Gray}
        \bf Model & \bf Method & \bf \#Bits & \bf Hella. & \bf PIQA & \bf ARC-c & \bf ARC-e & \bf Winogr. & \bf BoolQ & \bf OBQA & \bf Avg. \\ 
        \arrayrulecolor{black}
        \midrule

        LLaMA-1 13B & None & 16 & 59.9 & 79.2 & 46.5 & 77.4 & 72.8 & 78.0 & 33.2 & \bf 63.8 \\
        ~ & LoRA & 16 & 60.8 & 79.7 & 50.3 & 78.6 & 72.3 & 80.2 & 34.8 & \bf 65.2 \\
        \arrayrulecolor{lightgray}
        \cmidrule{2-11}
        ~ & GPTQ & 4 & 58.9 & 79.3 & 46.5 & 77.0 & 72.7 & 76.5 & 33.8 & \bf 63.5 \\
        ~ & OmniQ & 4 & 58.6 & 79.7 & 43.8 & 73.5 & 70.5 & 68.7 & 28.4 & \bf 60.4 \\
        ~ & LoftQ\textsuperscript{*} & 4\&16 & 60.6 & 79.0 & 48.3 & 77.7 & 72.9 & 76.0 & 35.0 & \bf 64.2 \\
        ~ & QLoRA\textsuperscript{*} & 4\&16 & 59.6 & 79.2 & 46.5 & 77.1 & 72.5 & 78.1 & 33.4 & \bf 63.8 \\
        ~ & QA-LoRA & 4 & 60.1 & 79.0 & 46.8 & 77.0 & 71.4 & 67.1 & 36.2 & \bf 62.5 \\
        ~ & QAT-LoRA & 4\&16 & 60.9 & 79.2 & 48.2 & 78.6 & 71.5 & 77.0 & 35.6 & \bf 64.4 \\
        ~ &   L4Q & 4 & 60.9 & 79.8 & 48.2 & 78.5 & 71.7 & 76.7 & 35.4 &   \bf 64.5 \\
        \arrayrulecolor{lightgray}
        \cmidrule{2-11}
        ~ & GPTQ & 3 & 57.3 & 77.3 & 42.6 & 73.0 & 71.0 & 74.6 & 31.4 & \bf 61.0 \\
        ~ & OmniQ & 3 & 56.8 & 77.2 & 39.9 & 72.7 & 68.5 & 67.0 & 29.8 & \bf 58.9 \\
        ~ & LoftQ\textsuperscript{*} & 3\&16 & 47.8 & 72.1 & 37.6 & 70.8 & 58.3 & 65.5 & 25.8 & \bf 54.0 \\
        ~ & QLoRA\textsuperscript{*} & 3\&16 & 56.6 & 77.8 & 43.9 & 75.1 & 70.8 & 73.5 & 31.6 & \bf 61.3 \\
        ~ & QA-LoRA & 3 & 57.7 & 78.0 & 44.7 & 75.3 & 71.2 & 68.6 & 32.4 & \bf 61.1 \\
        ~ & QAT-LoRA & 3\&16 & 59.1 & 78.1 & 46.3 & 77.0 & 70.8 & 74.7 & 36.2 & \bf 63.2 \\
        ~ &   L4Q & 3 & 58.9 & 78.4 & 45.8 & 77.4 & 70.2 & 77.7 & 35.2 &   \bf 63.4 \\
        \arrayrulecolor{black}
        \midrule

        LLaMA-1 33B & None & 16 & 63.3 & 81.0 & 52.8 & 80.4 & 75.9 & 82.6 & 36.0 & \bf 67.4 \\
        ~ & LoRA & 16 & 64.1 & 81.3 & 53.7 & 81.6 & 75.5 & 84.0 & 37.6 & \bf 68.3 \\
        \arrayrulecolor{lightgray}
        \cmidrule{2-11}
        ~ & GPTQ & 4 & 61.8 & 80.5 & 49.1 & 78.9 & 73.6 & 82.2 & 33.6 & \bf 65.7 \\
        ~ & OmniQ & 4 & 62.3 & 80.0 & 48.5 & 75.8 & 73.9 & 69.1 & 31.0 & \bf 62.9 \\
        ~ & LoftQ\textsuperscript{*} & 4\&16 & 63.3 & 80.3 & 51.8 & 81.4 & 75.3 & 82.9 & 37.0 & \bf 67.4 \\
        ~ & QLoRA\textsuperscript{*} & 4\&16 & 62.3 & 80.2 & 50.2 & 79.5 & 74.9 & 81.0 & 35.4 & \bf 66.2 \\
        ~ & QA-LoRA & 4 & 62.8 & 80.3 & 50.1 & 79.5 & 75.1 & 73.2 & 36.4 & \bf 65.3 \\
        ~ & QAT-LoRA & 4\&16 & 62.3 & 81.3 & 53.0 & 81.6 & 74.9 & 82.7 & 35.4 & \bf 67.3 \\
        ~ &   L4Q & 4 & 63.9 & 81.0 & 53.0 & 81.3 & 75.0 & 82.8 & 35.8 &   \bf 67.5 \\
        \arrayrulecolor{lightgray}\cmidrule{2-11}
        ~ & GPTQ & 3 & 60.9 & 78.7 & 46.7 & 77.5 & 74.7 & 82.2 & 34.8 & \bf 65.1 \\ 
        ~ & OmniQ & 3 & 61.4 & 79.6 & 46.0 & 74.2 & 74.3 & 71.3 & 29.2 & \bf 62.3 \\ 
        ~ & LoftQ\textsuperscript{*} & 3\&16 & 46.1 & 74.9 & 38.9 & 73.9 & 58.3 & 63.5 & 28.2 & \bf 54.8 \\
        ~ & QLoRA\textsuperscript{*} & 3\&16 & 59.6 & 78.7 & 46.0 & 76.8 & 72.5 & 81.6 & 34.6 & \bf 64.3 \\
        ~ & QA-LoRA & 3 & 61.1 & 79.6 & 47.8 & 78.0 & 73.8 & 79.3 & 33.0 & \bf 64.6 \\
        ~ & QAT-LoRA & 3\&16 & 63.3 & 81.0 & 52.8 & 81.3 & 75.5 & 82.8 & 35.4 & \bf 67.4 \\
        ~ &   L4Q & 3 & 63.0 & 81.0 & 52.6 & 81.4 & 75.5 & 82.8 & 35.4 &   \bf 67.4 \\
        \arrayrulecolor{black}
        \midrule
        
        LLaMA-2 7B & None & 16 & 57.1 & 78.1 & 43.4 & 76.3 & 69.1 & 77.7 & 31.4 & \bf 61.9 \\
        ~ & LoRA & 16 & 57.9 & 78.9 & 48.0 & 77.4 & 70.3 & 75.8 & 34.8 & \bf 63.3 \\
        \arrayrulecolor{lightgray}
        \cmidrule{2-11}
        ~ & GPTQ & 4 & 56.0 & 77.5 & 42.2 & 75.0 & 68.2 & 76.4 & 29.8 & \bf 60.7 \\
        ~ & OmniQ & 4 & 56.0 & 77.7 & 41.3 & 69.9 & 67.8 & 73.5 & 30.2 & \bf 59.5 \\
        ~ & LoftQ\textsuperscript{*} & 4\&16 & 57.0 & 78.0 & 43.3 & 76.3 & 69.2 & 76.8 & 31.4 & \bf 61.7 \\
        ~ & QLoRA\textsuperscript{*} & 4\&16 & 56.6 & 77.8 & 43.3 & 75.2 & 69.1 & 75.3 & 31.8 & \bf 61.3 \\
        ~ & QA-LoRA & 4 & 56.4 & 79.3 & 73.3 & 39.2 & 71.8 & 75.5 & 31.4 & \bf 61.0 \\
        ~ & QAT-LoRA & 4\&16 & 56.6 & 77.7 & 43.7 & 75.6 & 69.5 & 77.7 & 32.6 & \bf 61.9 \\
        ~ & L4Q & 4 & 57.2 & 78.8 & 47.1 & 76.9 & 70.2 & 80.4 & 34.8 &   \bf 63.6 \\
        \arrayrulecolor{lightgray}
        \cmidrule{2-11}
        ~ & GPTQ & 3 & 53.1 & 76.2 & 35.8 & 70.3 & 67.7 & 72.4 & 27.6 & \bf 57.6 \\
        ~ & OmniQ & 3 & 54.6 & 76.4 & 37.5 & 67.6 & 66.1 & 71.9 & 31.0 & \bf 57.9 \\
        ~ & LoftQ\textsuperscript{*} & 3\&16 & 27.1 & 55.7 & 19.0 & 31.1 & 48.8 & 48.1 & 12.8 & \bf 34.7 \\
        ~ & QLoRA\textsuperscript{*} & 3\&16 & 52.4 & 75.9 & 37.6 & 69.9 & 65.6 & 74.1 & 27.4 & \bf 57.6 \\
        ~ & QA-LoRA & 3 & 56.5 & 77.8 & 42.3 & 74.7 & 68.0 & 30.8 & 43.8 & \bf 56.3 \\
        ~ & QAT-LoRA & 3\&16 & 52.0 & 75.2 & 39.3 & 71.1 & 65.1 & 69.9 & 29.3 & \bf 57.4 \\
        ~ & L4Q & 3 & 55.5 & 77.3 & 42.8 & 73.8 & 68.8 & 77.2 & 34.0 &   \bf 61.3 \\
        \arrayrulecolor{black}
        \midrule
        LLaMA-2 13B & None & 16 & 60.1 & 79.1 & 48.5 & 79.4 & 72.2 & 80.6 & 35.2 & \bf 65.0 \\
        ~ & LoRA & 16 & 61.2 & 79.4 & 53.0 & 79.8 & 73.2 & 81.4 & 37.4 & \bf 66.5 \\
        \arrayrulecolor{lightgray}
        \cmidrule{2-11}
        ~ & GPTQ & 4 & 59.5 & 78.3 & 47.3 & 78.7 & 72.1 & 80.9 & 34.2 & \bf 64.4 \\
        ~ & OmniQ & 4 & 59.0 & 78.1 & 43.7 & 71.3 & 68.7 & 66.6 & 32.0 & \bf 59.9 \\
        ~ & LoftQ\textsuperscript{*} & 4\&16 & 60.0 & 79.3 & 48.1 & 79.7 & 71.9 & 80.7 & 34.8 & \bf 64.9 \\
        ~ & QLoRA\textsuperscript{*} & 4\&16 & 59.6 & 78.4 & 46.6 & 77.9 & 72.2 & 79.2 & 33.8 & \bf 64.0 \\
        ~ & QA-LoRA & 4 & 59.4 & 78.5 & 79.1 & 46.9 & 72.3 & 80.7 & 34.4 & \bf 64.5 \\
        ~ & QAT-LoRA & 4\&16 & 59.5 & 78.8 & 48.4 & 79.2 & 71.5 & 80.9 & 34.4 & \bf 64.7 \\
        ~ & L4Q & 4 & 60.9 & 80.1 & 51.2 & 79.7 & 71.0 & 82.2 & 35.8 &   \bf 65.8 \\
        \arrayrulecolor{lightgray}
        \cmidrule{2-11}
        ~ & GPTQ & 3 & 57.3 & 77.2 & 43.5 & 76.1 & 69.9 & 74.0 & 34.0 & \bf 61.7 \\
        ~ & OmniQ & 3 & 57.8 & 78.2 & 42.0 & 72.3 & 68.0 & 69.9 & 31.2 & \bf 59.9 \\
        ~ & LoftQ\textsuperscript{*} & 3\&16 & 28.7 & 60.6 & 19.5 & 45.3 & 50.7 & 55.1 & 15.2 & \bf 39.3 \\
        ~ & QLoRA\textsuperscript{*} & 3\&16 & 57.8 & 77.9 & 44.3 & 76.7 & 70.0 & 78.1 & 32.6 & \bf 62.5 \\
        ~ & QA-LoRA & 3 & 57.3 & 77.2 & 76.0 & 43.4 & 70.1 & 73.7 & 34.0 & \bf 61.7 \\
        ~ & QAT-LoRA & 3\&16 & 55.8 & 77.1 & 67.6 & 76.0 & 67.6 & 75.1 & 30.8 & \bf 64.3 \\
        ~ & L4Q & 3 & 59.3 & 78.7 & 51.2 & 78.5 & 70.6 & 79.9 & 37.4 &  \bf 65.1 \\
        \arrayrulecolor{black}
        \midrule

        Mistral-v0.1 7B & None & 16 & 61.2 & 80.6 & 50.4 & 80.9 & 73.9 & 83.6 & 32.6 & \bf 66.2 \\ 
        ~ & LoRA & 16 & 61.2 & 82.1 & 50.3 & 80.9 & 74.0 & 83.7 & 32.6 & \bf 66.4 \\
        \arrayrulecolor{lightgray}
        \cmidrule{2-11}
        ~ & GPTQ & 4 & 59.8 & 82.3 & 46.9 & 79.4 & 73.5 & 84.4 & 30.6 & \bf 65.3 \\ 
        ~ & OmniQ & 4 & 60.7 & 79.9 & 47.1 & 78.2 & 73.6 & 82.5 & 31.2 & \bf 64.7 \\ 
        ~ & LoftQ\textsuperscript{*} & 4\&16 & 54.2 & 78.0 & 44.5 & 77.6 & 67.5 & 75.1 & 27.8 & \bf 60.7 \\ 
        ~ & QLoRA\textsuperscript{*} & 4\&16 & 60.8 & 82.1 & 50.5 & 80.4 & 73.2 & 81.8 & 32.0 & \bf 65.8 \\ 
        ~ & QA-LoRA & 4 & 60.6 & 81.7 & 49.0 & 79.5 & 73.3 & 81.8 & 32.0 & \bf 65.4 \\ 
        ~ & QAT-LoRA & 4\&16 & 60.3 & 80.0 & 46.8 & 78.6 & 73.4 & 82.6 & 29.8 & \bf 64.5 \\ 
        ~ & L4Q & 4 & 60.3 & 81.6 & 50.9 & 80.4 & 72.4 & 84.8 & 32.0 & \bf 66.1 \\ 
        \arrayrulecolor{lightgray}
        \cmidrule{2-11}
        ~ & GPTQ & 3 & 57.3 & 79.5 & 43.5 & 75.8 & 70.6 & 78.0 & 27.8 & \bf 61.8 \\ 
        ~ & OmniQ & 3 & 58.7 & 79.1 & 43.4 & 75.2 & 69.8 & 72.3 & 31.0 & \bf 61.4 \\ 
        ~ & LoftQ\textsuperscript{*} & 3\&16 & 55.2 & 74.7 & 40.9 & 72.4 & 63.8 & 76.5 & 26.2 & \bf 58.5 \\ 
        ~ & QLoRA\textsuperscript{*} & 3\&16 & 57.5 & 80.3 & 46.6 & 76.7 & 69.8 & 80.7 & 29.6 & \bf 63.0 \\ 
        ~ & QA-LoRA & 3 & 57.4 & 78.7 & 43.9 & 75.7 & 70.9 & 80.9 & 28.4 & \bf 62.3 \\ 
        ~ & QAT-LoRA & 3\&16 & 56.8 & 79.7 & 40.9 & 74.6 & 70.3 & 79.5 & 29.4 & \bf 61.6 \\ 
        ~ & L4Q & 3 & 57.5 & 80.2 & 47.7 & 78.0 & 66.5 & 83.8 & 28.4 & \bf 63.1 \\ 
        
        \arrayrulecolor{black}
        \bottomrule
    \end{tabular}
    }
    \label{table:csqa_b}
\end{table}

\newpage
\renewcommand{\thetable}{14a}
\begin{table}[!ht]
    \caption{MMLU benchmark result. The numbers represent accuracy(\%) of each task.}
    \label{table:mmlu_a}
    \centering
    \renewcommand{\arraystretch}{0.5}
    \resizebox{\textwidth}{!}{
    \begin{tabular}{l l l | ccccc ccccc}
    \toprule
    \rowcolor{Gray}
    ~ & ~ & ~ & \multicolumn{5}{c|}{\bf 0-shot} & \multicolumn{5}{c}{\bf 5-shot} \\
    \rowcolor{Gray}
    \bf Model & \bf Method & \bf \#Bits & \bf Human. & \bf STEM & \bf Social & \bf Others & \bf Avg. & \bf Human. & \bf STEM & \bf Social & \bf Others & \bf Avg. \\
    \midrule
    
    LLaMA-1 7B & None & 16 & 32.9 & 26.9 & 32.1 & 37.3 & \bf 32.5 & 33.9 & 30.6 & 38.2 & 38.2 & \bf 35.1 \\
    ~ & LoRA & 16 & 36.1 & 31.5 & 36.9 & 40.6 & \bf 36.3 & 34.4 & 30.3 & 39.9 & 43.1 & \bf 36.7 \\
    \arrayrulecolor{lightgray}\cmidrule{2-13}
    ~ & GPTQ & 4 & 28.4 & 27.1 & 27.0 & 30.4 & \bf 28.3 & 31.5 & 30.4 & 33.7 & 35.7 & \bf 32.7 \\
    ~ & OmniQ & 4 & 31.1 & 26.7 & 29.8 & 35.5 & \bf 30.9 & 31.1 & 29.8 & 35.5 & 37.5 & \bf 33.3 \\
    ~ & LoftQ\textsuperscript{*} & 4\&16 & 32.3 & 30.0 & 32.7 & 37.3 & \bf 33.0 & 33.6 & 30.7 & 37.2 & 39.0 & \bf 35.1 \\
    ~ & QLoRA\textsuperscript{*} & 4\&16 & 33.1 & 27.1 & 33.1 & 37.5 & \bf 32.8 & 32.3 & 29.0 & 35.4 & 38.0 & \bf 33.6 \\
    ~ & QA-LoRA & 4 & 33.5 & 29.5 & 37.5 & 37.9 & \bf 34.5 & 34.1 & 31.2 & 38.5 & 39.0 & \bf 35.6 \\
    ~ & QAT-LoRA & 4\&16 & 33.5 & 29.5 & 34.6 & 37.4 & \bf 33.8 & 32.2 & 32.4 & 35.6 & 39.9 & \bf 34.8 \\
    ~ &   L4Q & 4 & 32.4 & 32.1 & 36.7 & 39.4 & \bf 34.9 & 34.2 & 30.7 & 38.4 & 39.8 & \bf 35.7 \\
    \arrayrulecolor{lightgray}\cmidrule{2-13}
    ~ & GPTQ & 3 & 25.0 & 22.5 & 22.0 & 24.5 & \bf 23.7 & 25.9 & 25.7 & 28.2 & 29.7 & \bf 27.3 \\
    ~ & OmniQ & 3 & 27.8 & 29.7 & 26.8 & 32.2 & \bf 29.0 & 31.6 & 32.1 & 33.7 & 29.7 & \bf 31.6 \\
    ~ & LoftQ\textsuperscript{*} & 3\&16 & 24.3 & 21.7 & 22.4 & 24.5 & \bf 23.4 & 23.4 & 23.2 & 21.9 & 23.5 & \bf 23.1 \\
    ~ & QLoRA\textsuperscript{*} & 3\&16 & 27.8 & 27.1 & 26.6 & 29.1 & \bf 27.7 & 30.5 & 28.6 & 32.1 & 34.9 & \bf 31.5 \\
    ~ & QA-LoRA & 3 & 28.9 & 27.1 & 25.8 & 29.6 & \bf 28.0 & 29.1 & 26.6 & 29.7 & 31.1 & \bf 29.1 \\
    ~ & QAT-LoRA & 3\&16 & 28.2 & 29.7 & 32.0 & 33.7 & \bf 30.6 & 29.8 & 29.2 & 32.2 & 34.6 & \bf 31.5 \\
    ~ &   L4Q & 3 & 29.5 & 27.8 & 32.1 & 33.3 & \bf 30.6 & 29.3 & 31.0 & 33.5 & 30.4 & \bf 31.8 \\
    \arrayrulecolor{black}\midrule
    
    LLaMA-1 13B & None & 16 & 41.0 & 36.5 & 49.3 & 48.6 & \bf 43.6 & 43.8 & 35.3 & 52.7 & 54.2 & \bf 46.3 \\
    ~ & LoRA & 16 & 42.4 & 34.0 & 49.4 & 51.9 & \bf 44.3 & 45.0 & 36.4 & 54.1 & 53.1 & \bf 47.0 \\
    \arrayrulecolor{lightgray}\cmidrule{2-13}
    ~ & GPTQ & 4 & 33.5 & 34.5 & 44.9 & 44.9 & \bf 40.1 & 43.1 & 35.9 & 52.8 & 51.9 & \bf 45.7 \\
    ~ & OmniQ & 4 & 39.8 & 35.1 & 48.6 & 48.1 & \bf 42.6 & 43.1 & 35.7 & 52.5 & 52.3 & \bf 45.7 \\
    ~ & LoftQ\textsuperscript{*} & 4\&16 & 39.0 & 34.8 & 47.8 & 48.5 & \bf 42.4 & 43.4 & 34.3 & 52.3 & 52.1 & \bf 45.4 \\
    ~ & QLoRA\textsuperscript{*} & 4\&16 & 39.0 & 35.7 & 47.5 & 47.2 & \bf 42.1 & 43.8 & 35.3 & 52.0 & 52.8 & \bf 45.9 \\
    ~ & QA-LoRA & 4 & 35.3 & 35.2 & 47.7 & 47.9 & \bf 42.4 & 43.0 & 34.6 & 53.0 & 53.6 & \bf 45.8 \\
    ~ & QAT-LoRA & 4\&16 & 39.8 & 33.9 & 46.9 & 48.3 & \bf 42.0 & 43.2 & 35.0 & 51.8 & 52.8 & \bf 45.5 \\
    ~ &   L4Q & 4 & 40.5 & 35.3 & 49.5 & 48.5 & \bf 43.2 & 43.4 & 36.1 & 52.3 & 53.2 & \bf 46.0 \\
    \arrayrulecolor{lightgray}\cmidrule{2-13}
    ~ & GPTQ & 3 & 32.1 & 27.2 & 36.3 & 36.8 & \bf 33.1 & 36.0 & 29.8 & 42.2 & 45.6 & \bf 38.2\\
    ~ & OmniQ & 3 & 32.6 & 28.2 & 37.1 & 41.8 & \bf 34.8 & 39.1 & 33.1 & 47.5 & 47.6 & \bf 41.6 \\
    ~ & LoftQ\textsuperscript{*} & 3\&16 & 25.1 & 24.7 & 24.0 & 26.1 & \bf 25.0 & 24.9 & 27.4 & 24.1 & 25.0 & \bf 25.3 \\
    ~ & QLoRA\textsuperscript{*} & 3\&16 & 33.3 & 31.3 & 38.5 & 42.3 & \bf 36.1 & 36.8 & 32.4 & 46.6 & 47.0 & \bf 40.4 \\
    ~ & QA-LoRA & 3 & 35.9 & 28.4 & 42.4 & 43.5 & \bf 37.5 & 34.8 & 31.5 & 43.0 & 44.8 & \bf 38.2 \\
    ~ & QAT-LoRA & 3\&16 & 37.6 & 31.6 & 41.9 & 44.3 & \bf 38.8 & 38.0 & 34.1 & 44.5 & 47.9 & \bf 40.9 \\
    ~ &   L4Q & 3 & 38.5 & 33.2 & 44.7 & 47.2 & \bf 40.7 & 39.3 & 34.0 & 46.6 & 48.4 & \bf 41.8 \\
    \arrayrulecolor{black}\midrule
    
    LLaMA-1 33B & None & 16 & 51.0 & 40.1 & 62.2 & 59.4 & \bf 53.0 & 54.4 & 44.7 & 65.4 & 61.6 & \bf 56.4 \\
    ~ & LoRA & 16 & 49.2 & 41.3 & 61.4 & 58.7 & \bf 54.4 &  55.2 & 46.1 & 66.4 & 63.3 & \bf 57.6\\
    \arrayrulecolor{lightgray}\cmidrule{2-13}
    ~ & GPTQ & 4 & 49.4 & 39.6 & 59.1 & 58.1 & \bf 51.4 & 52.5 & 45.1 & 64.2 & 62.2 & \bf 55.7 \\
    ~ & OmniQ & 4 & 48.5 & 40.3 & 61.3 & 59.1 & \bf 52.0 & 53.4 & 44.8 & 64.7 & 61.1 & \bf 55.8 \\
    ~ & LoftQ\textsuperscript{*} & 4\&16 & 49.2 & 40.2 & 60.8 & 58.0 & \bf 51.8 & 54.6 & 44.5 & 65.2 & 61.5 & \bf 56.4 \\
    ~ & QLoRA\textsuperscript{*} & 4\&16 & 48.5 & 39.0 & 59.7 & 57.8 & \bf 51.0 & 54.5 & 44.2 & 63.4 & 60.5 & \bf 55.6 \\
    ~ & QA-LoRA & 4 & 45.2 & 39.7 & 56.6 & 55.5 & \bf 48.9 & 52.7 & 43.5 & 63.4 & 61.0 & \bf 55.0 \\
    ~ & QAT-LoRA & 4\&16 & 50.2 & 39.8 & 60.9 & 58.9 & \bf 52.3 & 55.0 & 45.5 & 65.1 & 61.8 & \bf 56.7 \\
    ~ &   L4Q & 4 & 50.8 & 42.1 & 61.5 & 59.4 & \bf 53.3 & 53.5 & 46.6 & 66.1 & 61.8 & \bf 56.7 \\
    \arrayrulecolor{lightgray}\cmidrule{2-13}
    ~ &  GPTQ & 3 & 47.0 & 39.0 & 57.9 & 57.3 & \bf 50.0 & 49.4 & 42.3 & 59.2 & 57.5 & \bf 51.9 \\ 
    ~ & OmniQ & 3 & 46.5 & 40.0 & 59.0 & 56.7 & \bf 50.2 & 46.5 & 43.8 & 60.0 & 60.2 & \bf 52.4 \\ 
    ~ & LoftQ\textsuperscript{*} & 3\&16 & 24.7 & 24.0 & 23.2 & 26.4 & \bf 24.6 & 24.3 & 23.2 & 22.9 & 25.6 & \bf 24.0 \\
    ~ & QLoRA\textsuperscript{*} & 3\&16 & 41.8 & 34.6 & 55.2 & 52.3 & \bf 45.6 & 46.4 & 40.9 & 57.9 & 56.7 & \bf 50.1 \\
    ~ & QA-LoRA & 3 & 41.5 & 37.2 & 54.4 & 53.1 & \bf 46.1 & 45.4 & 39.9 & 55.6 & 55.0 & \bf 48.7 \\
    ~ & QAT-LoRA & 3\&16 & 47.7 & 38.9 & 58.0 & 56.6 & \bf 50.1 & 46.8 & 41.2 & 58.2 & 57.5 & \bf 50.6 \\
    ~ &   L4Q & 3 & 46.3 & 41.0 & 58.8 & 57.4 & \bf 50.5 & 50.4 & 42.9 & 61.0 & 59.1 & \bf 53.1 \\
    \arrayrulecolor{black}\midrule
    
    LLaMA-2 7B & None & 16 & 39.3 & 34.0 & 47.9 & 46.0 & \bf 41.6 & 42.8 & 37.0 & 50.6 & 52.2 & \bf 45.4 \\
    ~ & LoRA & 16 & 41.0 & 34.6 & 50.8 & 50.2 & \bf 43.9 & 43.4 & 37.0 & 51.8 & 52.4 & \bf 46.0 \\
    \arrayrulecolor{lightgray}\cmidrule{2-13}
    ~ & GPTQ & 4 & 36.0 & 30.1 & 41.3 & 41.1 & \bf 37.1 & 40.9 & 33.9 & 48.9 & 48.6 & \bf 42.9 \\
    ~ & OmniQ & 4 & 37.7 & 34.6 & 47.2 & 45.7 & \bf 41.0 & 42.4 & 37.7 & 51.1 & 51.6 & \bf 45.4 \\
    ~ & LoftQ\textsuperscript{*} & 4\&16 & 36.7 & 31.3 & 42.9 & 43.6 & \bf 38.5 & 41.5 & 34.6 & 49.5 & 49.8 & \bf 43.7 \\
    ~ & QLoRA\textsuperscript{*} & 4\&16 & 37.3 & 31.5 & 43.3 & 42.7 & \bf 38.6 & 42.1 & 35.9 & 50.2 & 51.1 & \bf 44.6 \\
    ~ & QA-LoRA & 4 & 37.3 & 32.3 & 43.5 & 43.0 & \bf 38.9 & 42.0 & 35.7 & 49.6 & 50.8 & \bf 44.4 \\
    ~ & QAT-LoRA & 4\&16 & 36.5 & 32.4 & 40.6 & 42.6 & \bf 37.9 & 41.8 & 35.2 & 48.6 & 50.2 & \bf 43.8 \\
    ~ &   L4Q & 4 & 38.7 & 33.8 & 45.6 & 46.4 & \bf 40.9 & 42.9 & 37.7 & 50.5 & 51.9 & \bf 45.5 \\
    \arrayrulecolor{lightgray}\cmidrule{2-13}
    ~ & GPTQ & 3 & 28.9 & 28.5 & 35.3 & 33.7 & \bf 31.3 & 36.0 & 31.7 & 39.3 & 43.5 & \bf 37.5 \\
    ~ & OmniQ & 3 & 33.1 & 30.4 & 39.1 & 35.5 & \bf 34.3 & 34.1 & 32.4 & 41.6 & 44.4 & \bf 37.7 \\
    ~ & LoftQ\textsuperscript{*} & 3\&16 &  24.1 & 21.3 & 21.8 & 23.8 & \bf 22.9 & 23.7 & 26.1 & 22.4 & 24.9 & \bf 24.2 \\
    ~ & QLoRA\textsuperscript{*} & 3\&16 & 30.2 & 29.1 & 36.0 & 35.5 & \bf 32.5 & 35.4 & 32.5 & 40.5 & 42.7 & \bf 37.6 \\
    ~ & QA-LoRA & 3 & 28.9 & 27.8 & 34.7 & 33.7 & \bf 31.0 & 36.0 & 31.6 & 39.5 & 43.4 & \bf 37.5 \\
    ~ & QAT-LoRA & 3\&16 & 31.1 & 27.2 & 33.9 & 33.8 & \bf 31.5 & 34.2 & 31.2 & 39.9 & 42.7 & \bf 36.8 \\
    ~ &   L4Q & 3 & 31.0 & 32.7 & 38.6 & 39.2 & \bf 34.9 & 34.3 & 32.3 & 42.2 & 44.9 & \bf 38.0 \\
    \arrayrulecolor{black}\bottomrule
    \end{tabular}
    }
\end{table}

\renewcommand{\thetable}{14b}
\begin{table}[!ht]
    \caption{MMLU benchmark result. The numbers represent accuracy(\%) of each task.}
    \label{table:mmlu_b}
    \centering
    \renewcommand{\arraystretch}{0.5}
    \resizebox{\textwidth}{!}{
    \begin{tabular}{l l l | ccccc ccccc}
    \toprule
    \rowcolor{Gray}
    ~ & ~ & ~ & \multicolumn{5}{c|}{\bf 0-shot} & \multicolumn{5}{c}{\bf 5-shot} \\
    \rowcolor{Gray}
    \bf Model & \bf Method & \bf \#Bits & \bf Human. & \bf STEM & \bf Social & \bf Others & \bf Avg. & \bf Human. & \bf STEM & \bf Social & \bf Others & \bf Avg. \\
    \arrayrulecolor{black} \midrule
    LLaMA-2 13B & None & 16 & 47.8 & 42.3 & 60.5 & 59.4 & \bf 52.1 & 52.0 & 43.8 & 63.0 & 61.2 & \bf 54.8 \\
    ~ & LoRA & 16 & 48.8 & 42.4 & 60.9 & 59.2 & \bf 52.5 & 54.4 & 44.3 & 63.4 & 60.8 & \bf 55.7 \\
    \arrayrulecolor{lightgray}\cmidrule{2-13}
    ~ & GPTQ & 4 & 46.5 & 40.2 & 57.7 & 56.8 & \bf 50.0 & 52.3 & 43.1 & 62.7 & 61.5 & \bf 54.7 \\
    ~ & OmniQ & 4 & 47.8 & 41.9 & 60.1 & 58.9 & \bf 51.8 & 53.0 & 43.0 & 62.5 & 60.5 & \bf 54.7 \\
    ~ & LoftQ\textsuperscript{*} & 4\&16 & 47.2 & 42.0 & 60.4 & 58.9 & \bf 51.7 & 52.6 & 43.2 & 62.8 & 60.1 & \bf 54.5 \\
    ~ & QLoRA\textsuperscript{*} & 4\&16 & 46.9 & 40.9 & 58.8 & 57.6 & \bf 50.7 & 51.3 & 43.1 & 62.5 & 60.8 & \bf 54.2 \\
    ~ & QA-LoRA & 4 & 46.5 & 40.8 & 58.3 & 57.4 & \bf 50.4 & 51.6 & 42.5 & 62.3 & 60.7 & \bf 54.1 \\
    ~ & QAT-LoRA & 4\&16 & 47.5 & 41.0 & 58.8 & 56.8 & \bf 50.7 & 50.3 & 42.9 & 62.3 & 60.7 & \bf 53.8 \\
    ~ &   L4Q & 4 & 48.4 & 41.8 & 60.4 & 58.4 & \bf 51.9 & 53.6 & 44.3 & 62.7 & 60.5 & \bf 55.2 \\
    \arrayrulecolor{lightgray}\cmidrule{2-13}
    ~ & GPTQ & 3 & 43.5 & 37.3 & 53.6 & 51.8 & \bf 46.3 & 46.3 & 42.7& 57.3& 56.2& \bf 50.4 \\
    ~ & OmniQ & 3 & 42.3 & 38.9 & 54.5 & 51.3 & \bf 46.3 & 43.4 & 43.0 & 58.8 & 56.5 & \bf 50.2 \\
    ~ & LoftQ\textsuperscript{*} & 3\&16 & 24.2 & 21.7 & 23.8 & 23.7 & \bf 23.5 & 24.6 & 28.5 & 24.0 & 27.4 & \bf 26.0 \\
    ~ & QLoRA\textsuperscript{*} & 3\&16 & 43.9 & 38.3 & 53.9 & 52.2 & \bf 46.8 & 48.5 & 41.1 & 57.7 & 55.9 & \bf 50.6 \\
    ~ & QA-LoRA & 3 & 43.4 & 37.4 & 53.7 & 52.1 & \bf 46.4 & 47.5 & 41.5 & 56.3 & 55.3 & \bf 49.9 \\
    ~ & QAT-LoRA & 3\&16 & 42.5 & 37.3 & 53.0 & 52.1 & \bf 45.9 & 44.8 & 40.5 & 56.3 & 55.7 & \bf 48.9 \\
    ~ &   L4Q & 3 & 43.7 & 39.0 & 54.4 & 52.2 & \bf 47.1 & 46.6 & 39.8 & 58.4 & 56.7 & \bf 50.0 \\
\arrayrulecolor{black} \midrule

    Mistral-v0.1 7B & None & 16 & 54.1 & 51.2 & 70.5 & 67.8 & \bf 60.2 & 56.5 & 52.6 & 73.5 & 70.4 & \bf 62.6 \\ 
    ~ & LoRA & 16 & 54.5 & 51.4 & 70.9 & 68.2 & \bf 60.6 & 56.8 & 52.9 & 73.9 & 70.8 & \bf 62.9 \\ 
    \arrayrulecolor{lightgray}\cmidrule{2-13}
    ~ & GPTQ & 4 & 51.8 & 47.4 & 68.1 & 65.5 & \bf 57.6 & 55.9 & 50.4 & 71.6 & 68.1 & \bf 61.0 \\ 
    ~ & OmniQ & 4 & 52.4 & 49.2 & 69.0 & 65.7 & \bf 58.4 & 55.1 & 51.8 & 71.2 & 68.5 & \bf 61.0 \\ 
    ~ & LoftQ\textsuperscript{*} & 4\&16 & 38.4 & 40.8 & 53.5 & 51.1 & \bf 45.2 & 39.9 & 41.6 & 53.4 & 50.8 & \bf 45.7 \\ 
    ~ & QLoRA\textsuperscript{*} & 4\&16 & 52.6 & 49.2 & 69.6 & 66.0 & \bf 58.7 & 55.7 & 51.7 & 71.2 & 68.0 & \bf 61.1 \\ 
    ~ & QA-LoRA & 4 & 51.4 & 46.4 & 66.0 & 64.1 & \bf 56.5 & 56.1 & 51.4 & 72.0 & 67.6 & \bf 61.2 \\ 
    ~ & QAT-LoRA & 4\&16 & 52.1 & 48.8 & 69.5 & 67.7 & \bf 58.8 & 54.1 & 50.5 & 71.0 & 68.0 & \bf 60.2 \\ 
    ~ & L4Q & 4 & 52.6 & 48.9 & 69.9 & 67.1 & \bf 59.0 & 56.3 & 51.1 & 71.7 & 68.8 & \bf 61.4 \\ 
    \arrayrulecolor{lightgray}\cmidrule{2-13}
    ~ & GPTQ & 3 & 45.0 & 42.4 & 59.7 & 57.2 & \bf 50.5 & 43.0 & 43.0 & 57.4 & 57.8 & \bf 49.6 \\ 
    ~ & OmniQ & 3 & 49.4 & 44.9 & 63.4 & 61.6 & \bf 54.3 & 50.0 & 47.0 & 66.1 & 63.0 & \bf 55.9 \\ 
    ~ & LoftQ\textsuperscript{*} & 3\&16 & 32.2 & 33.3 & 38.0 & 41.3 & \bf 35.8 & 34.6 & 34.1 & 39.6 & 41.1 & \bf 37.1 \\ 
    ~ & QLoRA\textsuperscript{*} & 3\&16 & 46.9 & 43.4 & 60.6 & 60.2 & \bf 52.2 & 47.6 & 44.8 & 61.6 & 62.9 & \bf 53.6 \\ 
    ~ & QA-LoRA & 3 & 45.8 & 42.9 & 60.8 & 54.5 & \bf 50.5 & 48.8 & 43.6 & 59.2 & 56.4 & \bf 51.7 \\ 
    ~ & QAT-LoRA & 3\&16 & 47.6 & 43.0 & 61.3 & 59.9 & \bf 52.4 & 48.5 & 46.3 & 63.6 & 60.2 & \bf 54.0 \\ 
    ~ & L4Q & 3 & 49.9 & 43.8 & 63.0 & 63.0 & \bf 54.5 & 51.1 & 46.1 & 66.4 & 62.5 & \bf 56.2 \\ 
    \arrayrulecolor{black}\bottomrule
    \end{tabular}
    }
\end{table}

\end{document}